%% file: 0.main.tex
\documentclass[sigconf]{acmart}

\usepackage{multirow}
\usepackage{subfigure}
\usepackage{bm}
\usepackage{multirow}
\usepackage{subfigure}
\usepackage[normalem]{ulem}
\useunder{\uline}{\ul}{}
\usepackage{subfigure}
\usepackage{algorithm}
\usepackage[noend]{algpseudocode}
\usepackage{enumitem}

\AtBeginDocument{%
  }

\setcopyright{acmlicensed}
\copyrightyear{2018}
\acmYear{2018}
\acmDOI{XXXXXXX.XXXXXXX}

\acmConference[Conference acronym 'XX]{Make sure to enter the correct
  conference title from your rights confirmation emai}{June 03--05,
  2018}{Woodstock, NY}

\acmISBN{978-1-4503-XXXX-X/18/06}

\begin{document}

\title{Harnessing the Synergy between LLM Agents and Knowledge Graphs for Urban Socioeconomic Prediction}

\author{Zhilun Zhou}
\orcid{0000-0002-8674-7513}
\affiliation{%
  \institution{Department of Electronic Engineering, BNRist, Tsinghua University}
  \city{Beijing}
  \country{China}  
}
\email{zzl22@mails.tsinghua.edu.cn}

\author{Jingyang Fan}
\affiliation{%
  \institution{Department of Electronic Engineering, BNRist, Tsinghua University}
  \city{Beijing}
  \country{China}
}
\email{siemprestr@gmail.com}

\author{Yu Liu}
\affiliation{%
  \institution{Department of Biomedical Engineering,
National University of Singapore,
}
    \country{Singapore}
}
\email{yu_liu@nus.edu.sg}

\author{Fengli Xu}
\authornote{Corresponding Author.}
\orcid{0000-0002-5720-4026}
\affiliation{%
  \institution{Department of Electronic Engineering, BNRist, Tsinghua University}
  \city{Beijing}
  \country{China}
}
\email{fenglixu@tsinghua.edu.cn}

\author{Depeng Jin}
\orcid{0000-0003-0419-5514}
\affiliation{%
  \institution{Department of Electronic Engineering, BNRist, Tsinghua University}
  \city{Beijing}
  \country{China}
}
\email{jindp@tsinghua.edu.cn}

\author{Yong Li}
\orcid{0000-0001-5617-1659}
\affiliation{%
  \institution{Department of Electronic Engineering, BNRist, Tsinghua University}
  \city{Beijing}
  \country{China}
}
\email{liyong07@tsinghua.edu.cn}

\begin{abstract}
    Socioeconomic prediction aims to leverage various urban data to predict the socioeconomic indicators of regions such as population and commercial activity level, which plays an important role in understanding urban regions and supporting decision-making.
    Existing studies leverage knowledge graphs (KG) to model heterogeneous urban data, and further apply graph representation learning methods for socioeconomic prediction. However, these approaches heavily rely on heuristic ideas and expertise to extract task-relevant knowledge from diverse data, which may not be optimal for specific tasks.
    Additionally, they tend to overlook the inherent relationships between different indicators, limiting the prediction accuracy.
    Motivated by the remarkable abilities of large language models (LLMs), in this work, we propose a synergistic framework of LLM agents and KG, which integrates the reasoning and representation learning on KG with LLM agents.
    We first construct an urban knowledge graph (UrbanKG) to model multi-sourced urban data and finetune an embedding language model (LM) to generate embeddings for KG entities with semantic information. Then we leverage the reasoning power of LLM to identify relevant meta-paths in the UrbanKG for each type of socioeconomic prediction task, and design a semantic-guided attention module for knowledge fusion with meta-paths. Moreover, we introduce a cross-task communication mechanism to further enhance performance by enabling knowledge sharing across tasks at both LLM agent and KG levels. On the one hand, the LLM agents for different tasks collaborate to generate more diverse and comprehensive meta-paths. On the other hand, the embeddings from different tasks are adaptively merged for better socioeconomic prediction.
    Experiments on two datasets demonstrate the effectiveness of the synergistic design between LLM and KG, providing insights for information sharing across socioeconomic prediction tasks.
    Our code is open-sourced \href{https://anonymous.4open.science/r/LLM_UrbanKG_Socioeconomic_Prediction-AB61/}{\textit{here}}\footnote{https://anonymous.4open.science/r/LLM\_UrbanKG\_Socioeconomic\_Prediction-AB61/} for reproducibility.
\end{abstract}





\maketitle

\input{1.Introduction}

\input{2.preliminaries}

\input{3.methods}

\input{4.experiments}

\input{5.related_work}

\input{6.conclusion}

\bibliographystyle{ACM-Reference-Format}
\bibliography{ref}

\newpage
\appendix

\input{7.Appendix}

\end{document}

%% file: 1.Introduction.tex
\section{Introduction}

Predicting an urban region's socioeconomic indicators, such as regional population and commercial activity, provides valuable insights into its development and plays a crucial role in scientific planning. With the proliferation of urban data, there has been a growing body of research focused on socioeconomic prediction using various urban data sources~\cite{wang2016crime,wang2017region,yang2017predicting,yao2018representing,dong2019predicting,xu2020attentional,zhang2021multi,wu2022multi,hou2022urban,luo2022urban,kim2022effective,zhou2023hierarchical}. 
Due to its promising ability to integrate heterogeneous data, Knowledge Graph (KG) has become an important approach to model multi-sourced urban data for socioeconomic prediction~\cite{zhou2023hierarchical}. 

Many existing studies leverage KG or other heterogeneous graphs to model the complex relationships within urban data and predict the socioeconomic indicators through graph representation learning methods. However, these approaches highly rely on heuristic ideas and expertise to extract knowledge related to the tasks, such as the construction of sub-graphs or definition of meta-structures~\cite{zhou2023hierarchical,kim2022effective}, which may be sub-optimal for different indicator prediction tasks. Moreover, the intrinsic correlations and potential for knowledge sharing across different socioeconomic prediction tasks are often overlooked, limiting the overall prediction accuracy.

The recent breakthrough of large language models (LLMs) provides an exciting new research direction. A line of recent studies focuses on leveraging the emergent reasoning ability of generative language models to mine the logical relations on KGs~\cite{feng2023knowledge,jiang2023structgpt,wang2024knowledge}, while some other studies aim to improve KG representation learning with large embedding language models~\cite{wang2023reasoning,wang2022language,xie2023lambdakg}. However, there remains a significant gap in exploring the potential synergistic effect between reasoning and representation learning in an integrative framework.

To address these limitations, we aim to utilize the following abilities of LLMs to better synergize LLM and KG, and improve socioeconomic prediction performance:
(1) \textbf{Latent embedding.} Language models are able to generate text embeddings with rich semantic information, which makes it possible to capture semantic information of entities and paths in KG and integrate it into representation learning models.
(2) \textbf{Explicit reasoning.} Moreover, LLMs also exhibit power general-purpose reasoning abilities~\cite{zhao2024large}. This capability may facilitate the automatic discoveries of task-relevant meta-structures, reducing the need for expert knowledge.
(3) \textbf{Multi-agent collaboration.} LLM agents can collaborate with other agents to solve complex tasks that are difficult for a single agent~\cite{li2023camel,hong2023metagpt,xiao2023chain}. Such ability enables LLM to collaboratively refine the knowledge extracted from KG, and transfer knowledge across different socioeconomic prediction tasks via semantic-rich natural language.

\begin{figure}[t]
    \centering
    \vspace{-10px}
    \includegraphics[width=.99\linewidth]{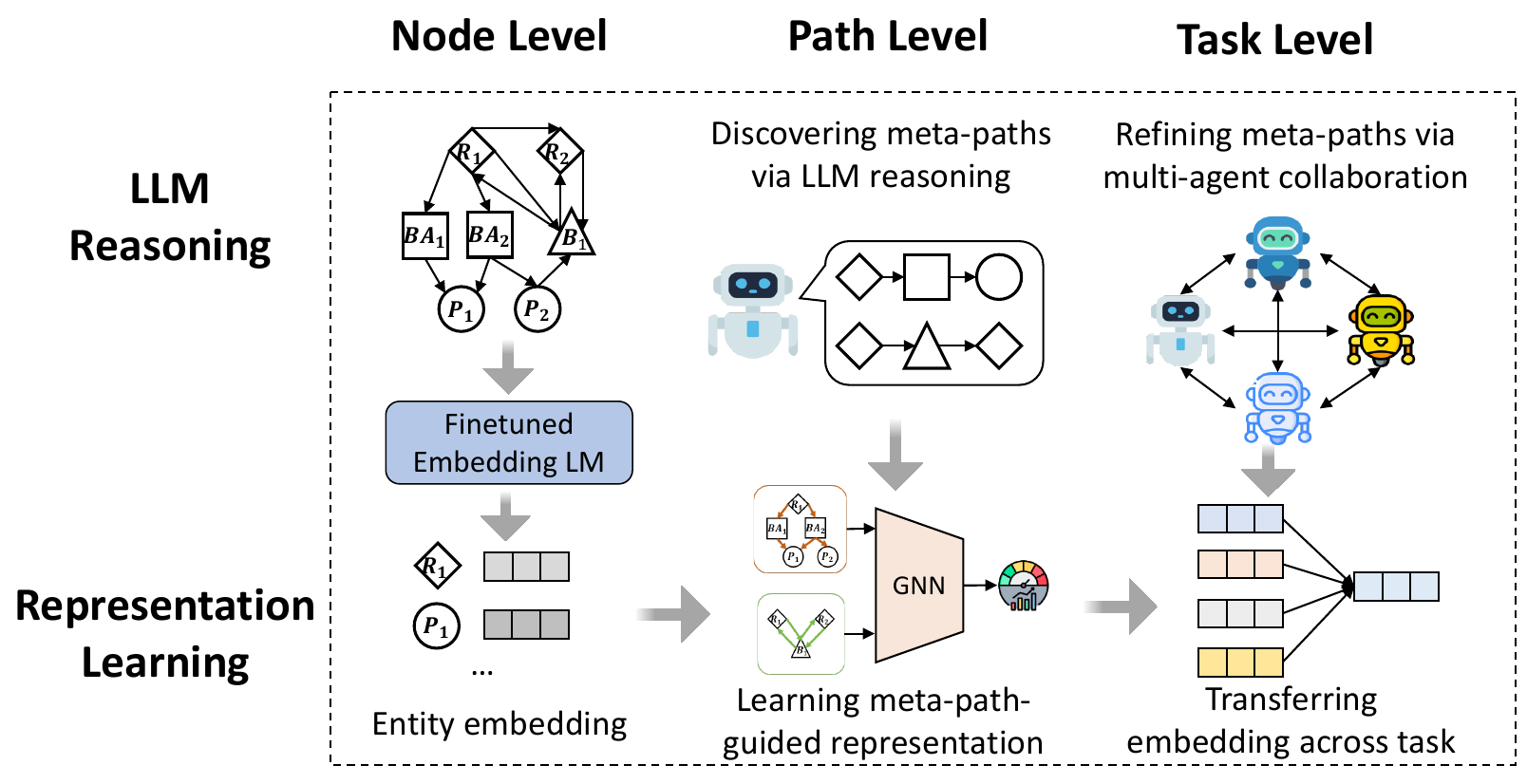}
    \vspace{-10px}
    \caption{The synergy between LLM reasoning and knowledge graph representation learning at different level.}
    \label{fig:concept}
    \vspace{-10px}
\end{figure}

In this work, we propose a method to \underline{S}ynergize \underline{L}LM \underline{A}gent and \underline{K}nowledge Graph learning model (SLAK) for socioeconomic prediction. Following previous studies~\cite{wang2021spatio,liu2021improving,liu2023knowsite,zhou2023hierarchical}, we construct an urban knowledge graph (UrbanKG) to integrate urban data comprehensively in a single graph. Then we propose a framework that combines reasoning and representation learning on UrbanKG with LLM agents, as shown in Figure~\ref{fig:concept}.
(1) At the node level, we leverage an embedding LM, finetuned on UrbanKG, to generate embeddings for entities with semantic information. 
(2) At the path level, we construct an LLM agent to automatically discover meta-paths in the UrbanKG in order to extract knowledge relevant to the specific socioeconomic indicator. We extract a sub-KG based on each meta-path from the UrbanKG. Moreover, we design a KG representation learning model to distil knowledge from each sub-KG, and adaptively fuse the knowledge based on semantic information of meta-paths.
(3) At the task level, we propose a cross-task communication mechanism to enable knowledge sharing across different socioeconomic prediction tasks. Specifically, we let the LLM agents collaborate to refine the meta-paths for each task, and propose a knowledge fusion module to adaptively merge the region embeddings from different tasks for better prediction.

Our contribution can be summarized as follows:
\begin{itemize}
    \item We propose a framework to synergize LLM agents and KG for socioeconomic prediction. We finetune an embedding LM to integrate semantic information into KG learning model, use the reasoning capability of LLM to find task-relevant meta-paths from UrbanKG, and design a cross-task communication mechanism to enable knowledge sharing across different tasks based on multi-agent collaboration.
    \item Our framework integrates two paradigms of KG applications, i.e., KG reasoning and representation learning, at the node level, path level, and task level, providing insights for KG-based applications.
    \item Extensive experiments on two city-scale datasets show that our model surpasses existing methods by 3.9-76.2\% in terms of $R^2$ on eight indicator prediction tasks, demonstrating the effectiveness of our synergistic model between LLM and KG. Several in-depth analyses further show the advantage of our model design.
\end{itemize}

%% file: 2.preliminaries.tex
\section{Preliminaries}
\label{sec:preliminaries}

In this section, we introduce some basic concepts, and then formally define the socioeconomic prediction problem. First, the object of prediction is urban region, which is defined as follows.

\subsection{Problem Statement}
\begin{definition}[\textbf{Urban Region}]
We define urban regions as non-overlapping irregular areas in a city, which are partitioned by main road networks, such as a block, denoted as $\mathcal{L}=\{L_1, L_2,\ldots, L_{N_L}\}$.
\end{definition}

\begin{definition}[\textbf{Socioeconomic Prediction Problem}]
Given a set of socioeconomic indicators $\mathcal{I}=\{I_1,\ldots, I_{N_I}\}$ of urban regions such as population, commercial activity, social activity and service quality, the socioeconomic prediction task aims to predict the value of these indicators for urban regions based on various urban data, i.e., learn a mapping function $f:\mathcal{L}\rightarrow \mathcal{V_I}$, where $\mathcal{V_I}$ is the range of indicator $I\in\mathcal{I}$.
\end{definition}

\subsection{Urban Knowledge Graph}
KG is a multi-relational graph structure, defined as $\mathcal{G}=\{\mathcal{E},\mathcal{R},\mathcal{F}\}$, where $\mathcal{E}$ is the entity set, $\mathcal{R}$ is the relation set, and $\mathcal{F}$ is the fact set. Each fact in $\mathcal{F}$ is denoted as a triplet $(h,r,t)$, representing a directional edge from the head entity $h\in\mathcal{E}$ to tail entity $t\in\mathcal{E}$ with relation type $r\in\mathcal{R}$. Due to its remarkable ability to represent heterogeneous data and integrate diverse knowledge, KG has been widely used in the study of urban computing~\cite{wang2021spatio,liu2021improving,liu2023knowsite,zhou2023hierarchical}.
Inspired by this, we construct an UrbanKG to integrate multi-sourced urban data~\cite{liu2023urbankg,zhou2023hierarchical}. The UrbanKG contains various elements in the city such as regions, POIs, POI categories, brands, business areas, and various relations between them. The details of UrbanKG are presented in Appendix~\ref{sec:KG_details}.

%% file: 3.methods.tex
\section{Methods}
\label{sec:methods}

\begin{figure*}[h]
    \centering
    \vspace*{-10px}
    \includegraphics[width=\linewidth]{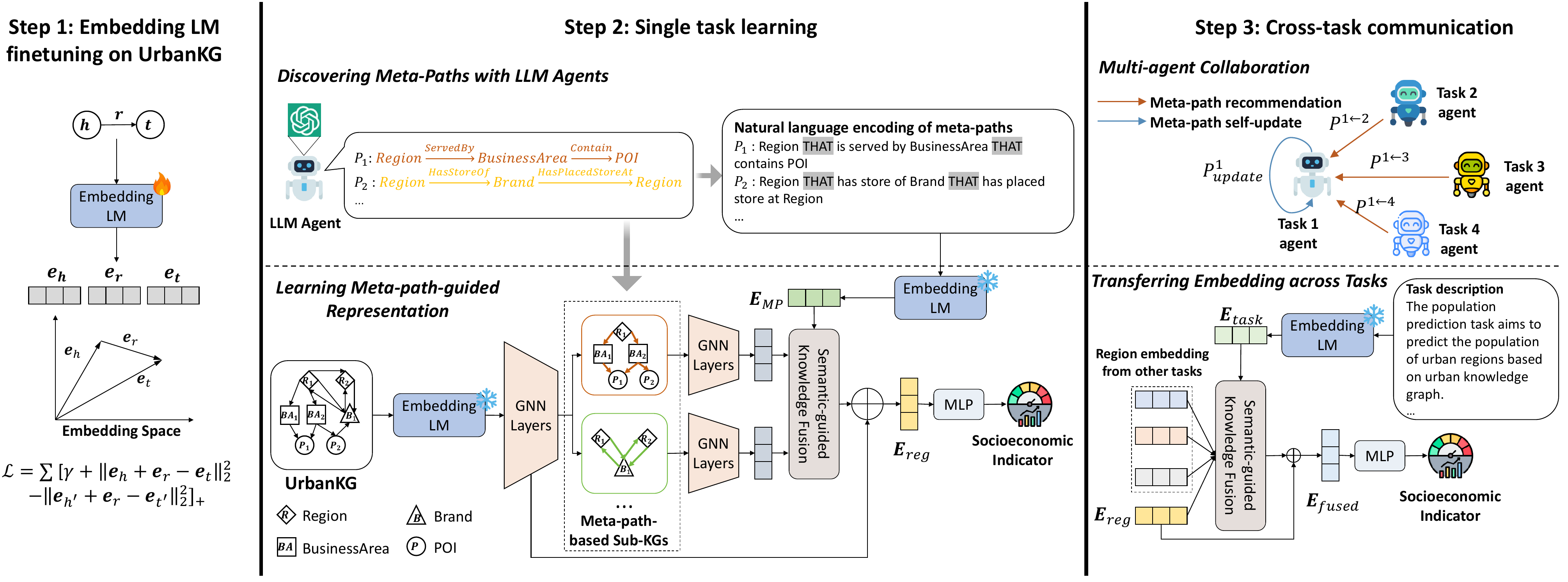}
    \vspace*{-10px}
    \caption{The overall framework of our proposed model SLAK.}
    \label{fig:framework}
    \vspace*{-10px}
\end{figure*}

\subsection{Framework Overview}

In this study, we propose a framework synergizing the capability of LLM agents and KG, as shown in Figure~\ref{fig:framework}. First, we finetune an embedding LM on UrbanKG and generate semantic embeddings for entities in UrbanKG, capturing the semantic information therein.
Next, we construct an LLM agent to extract task-relevant knowledge by prompting it to find meta-paths related to the socioeconomic prediction task from the UrbanKG, and extract a corresponding sub-KG based on each meta-path. The meta-path-based sub-KGs are fed into a KG learning model to distil knowledge from each sub-KG, and we further use the semantic embeddings of meta-paths to guide the knowledge fusion. The output region embeddings are used for socioeconomic prediction through a final MLP output layer.
Finally, we design a cross-task communication mechanism. Different indicator prediction tasks share knowledge at both the LLM agent level and the KG level. Specifically, the LLM agents recommend potential meta-paths for other tasks from diverse perspectives, and update their own meta-paths selection based on knowledge from other tasks. Moreover, we leverage embedding LM to obtain a semantic embedding for each task, which is used to adaptively fuse the region embeddings from different tasks.

\subsection{Semantic-enriched KG Embedding}
\label{sec:finetune}
The UrbanKG contains rich semantic information in its entities, relations, and paths. To extract such information and integrate it into prediction model, we leverage an embedding LM to obtain text embeddings. Here we choose the GTE-based model~\cite{li2023towards}, which is one of the top models on the text embedding leaderboard MTEB~\cite{muennighoff2022mteb}.
To align the embedding model with UrbanKG, we first finetune the model on a KG completion task. We leverage the scoring function of a classic KG representation learning model TransE~\cite{bordes2013translating}. Specifically, let $\bm{e}_h$, $\bm{e}_r$, and $\bm{e}_t$ denote the embeddings of head entity $h$, relation $r$ and tail entity $t$ obtained from the embedding model, the goal is that the sum of head entity embeddings and relation embeddings $\bm{e}_h + \bm{e}_r$ should be close to the tail entity embeddings $\bm{e}_t$.
Therefore, we optimize the following loss function:
\begin{equation}
\label{eqn:transe}
    \mathcal{L} = \sum_{(h,r,t)\in\mathcal{F}}[\gamma + ||\bm{e}_h + \bm{e}_r - \bm{e}_t||_2^2 - ||\bm{e}_{h'} + \bm{e}_r - \bm{e}_{t'}||_2^2]_{+},
\end{equation}
where $\gamma$ is a margin hyperparameter, $[x]_+=max\{x,0\}$, and $h'$, $t'$ are negative samples such that the triplets $(h',r,t)$ or $(h,r,t')$ does not exist in UrbanKG.
After finetuning, we use the model to obtain semantic embeddings for all entities in UrbanKG as the initial features in the following KG learning models.

\subsection{Single-task Learning}
\subsubsection{Meta-path Extraction}

The UrbanKG integrates various urban data comprehensively and incorporates knowledge from various domains, such as the spatial knowledge of geographical relationships between regions, the functional knowledge regarding POIs and POI categories, and mobility knowledge about population flow~\cite{zhou2023hierarchical}. It is crucial to identify relevant information that can aid in predicting specific socioeconomic indicators. Inspired by the ability of meta-paths in KG to capture certain semantic contexts, we aim to find the most relevant meta-paths that contribute to the prediction tasks.
We first provide the definition of meta-path as follows~\cite{sun2012mining}.
\begin{definition}[\textbf{Meta-path}]
A meta-path in KG can be represented in the form of $E_1\xrightarrow{R_1}E_2\xrightarrow{R_2}\ldots\xrightarrow{R_{l-1}}E_l$, denoting a path from entity $E_1, E_2,\ldots$ to $E_l$ through relations $R_1,\ldots,R_{l-1}$.
\end{definition}
Meta-paths provide rich semantic contexts in the UrbanKG. For example, the meta-path $Region\xrightarrow{Has}POI\xrightarrow{Competitive}POI\xrightarrow{LocateAt}Region$ captures the competitive relationships between POIs in two regions, which may affect the commercial activity of these regions.

However, given the large number of entities and relations in UrbanKG, it is not trivial to find helpful meta-paths. 
Here we draw inspiration from the emergent commonsense reasoning ability of LLMs. It has been demonstrated that LLM can identify important meta-structures in heterogeneous information networks through reasoning~\cite{chen2024large}.
Therefore, we construct an LLM agent to automatically find relevant meta-paths from UrbanKG.
To enable the agent to understand the UrbanKG, we input the schema of UrbanKG described in natural language. Then we prompt the agent to generate several potential meta-path schemes for the socioeconomic prediction task. 
Moreover, to ensure that LLM agents find valid meta-paths in UrbanKG, we provide all the valid triplets (including head entity type, relation type, tail entity type) in the prompt (Appendix Figure~\ref{fig:prompt_kginfo}), and restrict the content and format of LLM agent's output (Appendix Figure~\ref{fig:prompt_singletask}).

\subsubsection{KG Learning Model}
Following previous work~\cite{zhou2023hierarchical}, we employ a hierarchical KG learning model to learn region embeddings from UrbanKG based on the extracted meta-paths. 
First, we use GNN layers to extract global knowledge from the UrbanKG. Here we adopt a graph convolution model called R-GCN~\citep{schlichtkrull2018modeling}. Different from vanilla GCN, R-GCN aggregates information from neighboring nodes through each type of relation separately, which can better capture the structural information in KG. Specifically, let $\bm{e}_i^{(l)}$ denote the embedding of entity $e_i$ at the $l$-th R-GCN layer, the information aggregation function is:
\begin{equation}
    \bm{e}_i^{(l+1)}=\sigma(\sum_{r\in \mathcal{R}}\sum_{j\in \mathcal{N}_i^r}W_r^{(l)}\bm{e}_j^{(l)}+W_0^{(l)}\bm{e}_i^{(l)}),
\end{equation}
where $\bm{e}_i^{(l+1)}$ is the embedding at the $(l+1)$-th layer, $\mathcal{N}_i^r$ is the neighboring entities of $e_i$ with respect to relation $r$, and $W_r^{(l)}$, $W_0^{(l)}$ are learnable weight matrices. 
Note that the initial embeddings $\bm{e}_i^{(0)}$ are the generated by the finetuned embedding LM.
The output of R-GCN layers are the embeddings of all entities and relations in UrbanKG, which aims to capture the global knowledge in the UrbanKG.

Then we distil domain knowledge from the meta-paths. Given a meta-path scheme generated by LLM agent, we first extract all meta-paths with such scheme from the UrbanKG to construct a sub-KG, as shown in Figure~\ref{fig:framework}. The sub-KG preserves all the contextual knowledge defined by the meta-path scheme in UrbanKG. In practice, we use the graph database Neo4j\footnote{https://neo4j.com/} to store the UrbanKG and query it to get the meta-paths with Cypher language. 
Consequently, the output embeddings of R-GCN layers are used to initialize the embeddings of each sub-KG, and then fed into another R-GCN model, to obtain region embeddings from each meta-path.

\subsubsection{Semantic-guided Knowledge Fusion}
\label{sec:knowledge_fusion}
Different meta-paths have different importance in socioeconomic indicator prediction tasks. Therefore, we propose a semantic-guided knowledge fusion module to dynamically fuse the knowledge of different meta-paths.
We first obtain the natural language encoding of meta-paths using methods in~\cite{chen2024large}, which provides a description of the meta-path easy for LLM to understand. Specifically, we use the conjunction word \textit{THAT} to construct a nested clause for each meta-path, as shown in Figure~\ref{fig:framework}. The constructed sentences align with commonly-used English grammar, and thus are suitable for language models.
Moreover, we use the fine-tuned embedding LM (Section~\ref{sec:finetune}) to generate embeddings for these sentences, resulting in a $d_{LLM}=768$ dimensional semantic embedding for each meta-path.

Let $P_1,\ldots,P_{N_P}$ denote the meta-paths, and $\bm{E}_{MP}\in\mathbb{R}^{N_P\times d_{LLM}}$ are the semantic embeddings of meta-paths. We calculate the importance of meta-paths based on their semantic embeddings as:
\begin{equation}
\label{eqn:weight}
    w_{P_i}=\bm{q}^\top \tanh(\bm{W}\bm{E}_{MP,i}+\bm{b}),
\end{equation}
where $\bm{q}$ is the attention vector, $\bm{W}$ is the weight matrix and $\bm{b}$ is the bias vector. We normalize the weights with softmax function:
\begin{equation}
\label{eqn:softmax}
    \alpha_{P_i}=\frac{\exp(w_{P_i})}{\sum_{j=1}^{N_P}\exp(w_{P_j})}.
\end{equation}
Then, the output embedding of region $L_j$ is calculated as:
\begin{equation}
\label{eqn:attention_sum}
    \bm{e}_j = \sum_{i=1}^{N_P}\alpha_{P_i}\bm{e}_j^{P_i}.
\end{equation}
Finally, the above embeddings are added to the output of the R-GCN layers through a residual connection, and fed into an MLP layer for indicator prediction.
We train our model in a supervised manner and employ MSE loss to optimize it.


\algrenewcommand\algorithmicprocedure{\textbf{Round}}
\algnewcommand{\LineComment}[1]{\State \(\triangleright\) #1}
\begin{algorithm}
\caption{Training algorithm of our model}\label{alg:training}
\begin{algorithmic}[1]
\State Finetune embedding LM with Equation~\ref{eqn:transe} on UrbanKG
\State Generate embeddings $\bm{e}^{(0)}$ for all entities in UrbanKG
\Procedure{1: Single-task training}{}
    \For{Indicator $I_i\in\{I_1,\ldots,I_{N_I}\}$}
        \State Generate meta-paths $\mathcal{P}^{i}=\{P_1^{i},\ldots,P_{N_P}^{i}\}$ through LLM agent $A_i$
        \State Extract corresponding sub-KGs from UrbanKG
        \State Generate semantic embeddings of meta-paths $E_{MP}$
        \State Train the KG learning model to predict $I_i$ for many epochs with early-stopping
        \State Save the region embeddings $E_{reg}^{I_i}$ at the best epoch
    \EndFor
\EndProcedure
\Procedure{2: Cross-task communication}{}
    \LineComment{\textbf{Meta-paths update}}
    \For{ Indicator $I_i\in\{I_1,\ldots,I_{N_I}\}$}
        \State Agent $A_i$ updates its own meta-paths $\mathcal{P}_{update}^{i}=\{P_{1,update}^{i},\ldots,P_{N_P,update}^{i}\}$
        \State Agent $A_i$ recommends a meta-path to each other task $P^{j\leftarrow i}, j\neq i$
    \EndFor
    
    \LineComment{\textbf{Model training}}
    \For{Indicator $I_i\in\{I_1,\ldots,I_{N_I}\}$}
        \State Obtain the final meta-paths $\mathcal{P}_{new}^{i}=\mathcal{P}_{update}^{i}\cup\{P^{i\leftarrow j}|j\neq i\}$
        \State Train the model to predict $I_i$ with meta-paths $\mathcal{P}_{new}^{i}$ and embeddings $\{E_{reg}^{I_j}|j\neq i\}$
    \EndFor
\EndProcedure
\end{algorithmic}
\end{algorithm}

Since different socioeconomic indicators are inherently correlated, we further design a cross-task communication mechanism to enable knowledge sharing across different tasks regarding both the meta-paths finding and the embedding learning, as shown in Figure~\ref{fig:framework}. 
We conduct cross-task communication at both the agent level and KG level. For the agent-level communication, we leverage the multi-agent collaboration ability of LLMs to refine the meta-paths extracted from UrbanKG for each task. For the KG-level communication, we transfer the KG embeddings of regions across different socioeconomic prediction tasks.

Previous studies regarding multi-agent collaboration have shown that when assigned different roles, multiple LLM agents can think from different perspectives and communicate with each other to provide a more comprehensive and accurate answer to questions~\cite{lan2024stance,chan2023chateval}.
Inspired by this, we provide the agents with the previously found meta-paths of all tasks, and ask them to recommend a new meta-path to each of the other agents from their own task's perspective. We leverage the chain-of-thought prompting techniques~\cite{wei2022chain}, and ask the agent to consider the relationships between two tasks and provide the detailed thinking process. For example, the agent responsible for the population prediction task may think about how the population affects the commercial activity, and recommend a relevant meta-path to the commercial activity prediction task.
In addition, we ask each agent to update the previous meta-paths of its own tasks based on those from other tasks.
We combine the self-updated meta-paths as well as meta-paths recommended by other agents as the final meta-paths for each task.

Moreover, we transfer the region embeddings across tasks to further enhance the performance. We only modify the final layer of the KG learning model in Figure~\ref{fig:framework} by combining the current task embedding $E_{reg}$ with embeddings previously learned from other tasks.
Here we adopt the same semantic-guided knowledge fusion module as mentioned in Section~\ref{sec:knowledge_fusion}. We first construct a description of each task in natural language, and leverage the embedding LM to obtain the semantic embedding of each task, denoted as $E_{task}$. Then $E_{task}$ is used as the query of the attention module to adaptively fuse the embeddings using Equation~\ref{eqn:weight}-\ref{eqn:attention_sum}. The output of the semantic-guided attention module is added to $\bm{E}_{reg}$ of the current task through a residual connection, and the obtained region embeddings $\bm{E}_{fused}$ are finally used for socioeconomic prediction.

The overall training algorithm of our model is presented in Algorithm~\ref{alg:training}. Note that it consists of two rounds of training. In the first round, we train the model for each task separately, and save the final region embeddings $E_{reg}^{I_i}$. In the second round, the LLM agents communicate to generate new meta-paths $\mathcal{P}_{new}^{i}$ for each task, and we train the model for each task based on $\mathcal{P}_{new}^{i}$ and region embeddings from the previous round.

%% file: 4.experiments.tex
\section{Experiments}
\label{sec:experiments}

\begin{table}[h]
    \centering
    \caption{The basic information of two real-world datasets.}
    \vspace*{-10px}
    \resizebox{0.95\linewidth}{!}{
\begin{tabular}{cc|cc}
\toprule
\multicolumn{1}{l}{\textbf{}}                                                          & \textbf{City}        & Beijing                                                      & Shanghai                                                     \\ \hline
\multirow{2}{*}{\textbf{\begin{tabular}[c]{@{}c@{}}Basic \\ Info.\end{tabular}}} & \textbf{\#Regions}   & 523                                                          & 2032                                                         \\
                                                                                       & \textbf{Indicators}  & \multicolumn{2}{c}{\begin{tabular}[c]{@{}c@{}}population, commercial activity, \\ social activity,   service quality\end{tabular}} \\ \hline
\multirow{3}{*}{\textbf{\begin{tabular}[c]{@{}c@{}}UrbanKG \\ Statistics\end{tabular}}}   & \textbf{\#Entities}  & 23,754                                                       & 41,338                                                       \\
                                                                                       & \textbf{\#Relations} & 35                                                           & 35                                                           \\
                                                                                       & \textbf{\#Facts}     & 330,652                                                      & 589,850                                                      \\ 
\bottomrule
\end{tabular}
    }
    \label{tbl:datasets_info}
    \vspace*{-10px}
\end{table}

\begin{table*}[h]
    \centering
    \caption{Performance comparison with baselines on Beijing dataset. Best result are presented in bold, and the second best results are underlined.}
    \vspace*{-10px}
    \resizebox{0.9\textwidth}{!}{
\begin{tabular}{c|ccc|ccc|ccc|ccc}
\toprule
\multicolumn{1}{l|}{} & \multicolumn{3}{c|}{\textbf{Population}}         & \multicolumn{3}{c|}{\textbf{Commercial Activity}} & \multicolumn{3}{c|}{\textbf{Social Activity}}    & \multicolumn{3}{c}{\textbf{Service   Quality}}   \\
\textbf{Model}        & \textbf{MAE}   & \textbf{RMSE}  & \textbf{R$^2$} & \textbf{MAE}    & \textbf{RMSE}  & \textbf{R$^2$} & \textbf{MAE}   & \textbf{RMSE}  & \textbf{R$^2$} & \textbf{MAE}   & \textbf{RMSE}  & \textbf{R$^2$} \\ \hline
GAT                   & 0.676          & 0.916          & 0.305          & 1.293           & 1.712          & 0.270          & 2.164          & 3.125          & 0.115          & 0.706          & 1.056          & 0.058          \\
Metapath2vec          & 0.872          & 1.175          & 0.182          & 2.396           & 4.871          & 0.132          & 2.529          & 3.451          & 0.155          & 0.779          & 1.121          & 0.085          \\
MVURE                 & 0.707          & 0.920          & 0.299          & 1.051           & 1.388          & 0.521          & 1.907          & 2.631          & 0.372          & 0.686          & 1.022          & 0.119          \\
MGFN                  & 0.667          & 0.904          & 0.324          & 3.235           & 6.155          & 0.200          & 5.552          & 8.657          & 0.179          & 0.772          & 1.002          & 0.152          \\
HUGAT                 & 0.695          & 0.919          & 0.301          & 1.165           & 1.558          & 0.396          & 1.947          & 2.987          & 0.191          & 0.728          & 1.080          & 0.014          \\
HAFusion              & 0.708          & 0.965          & 0.229          & 1.207           & 1.590          & 0.371          & 1.712          & 2.562          & {\ul 0.405}    & 0.642          & 1.061          & 0.050          \\
HKGL                  & 0.660          & 0.886          & 0.350          & 1.000           & 1.238          & 0.619          & 1.864          & 2.618          & 0.379          & 0.752          & 1.031          & 0.102          \\
HKGL-trans            & 0.790          & 1.007          & 0.162          & 1.066           & 1.379          & 0.527          & 1.827          & 2.726          & 0.326          & 0.736          & 1.058          & 0.055          \\ \hline
SLAK-single           & {\ul 0.658}    & {\ul 0.884}    & {\ul 0.353}    & {\ul 0.960}     & {\ul 1.212}    & {\ul 0.634}    & \textbf{1.626} & {\ul 2.561}    & {\ul 0.405}    & {\ul 0.612}    & {\ul 0.936}    & {\ul 0.260}    \\
SLAK-comm             & \textbf{0.644} & \textbf{0.859} & \textbf{0.389} & \textbf{0.910}  & \textbf{1.180} & \textbf{0.654} & {\ul 1.657}    & \textbf{2.489} & \textbf{0.438} & \textbf{0.609} & \textbf{0.931} & \textbf{0.268} \\ \bottomrule
\end{tabular}
    }
    \label{tbl:main_result_beijing}
\end{table*}

\begin{table*}[h]
    \centering
    \caption{Performance comparison with baselines on Shanghai dataset. Best result are presented in bold, and the second best results are underlined.}
    \vspace*{-10px}
    \resizebox{0.9\textwidth}{!}{
\begin{tabular}{c|ccc|ccc|ccc|ccc}
\toprule
\multicolumn{1}{l|}{} & \multicolumn{3}{c|}{\textbf{Population}}         & \multicolumn{3}{c|}{\textbf{Commercial Activity}} & \multicolumn{3}{c|}{\textbf{Social Activity}}    & \multicolumn{3}{c}{\textbf{Service   Quality}}   \\
\textbf{Model}        & \textbf{MAE}   & \textbf{RMSE}  & \textbf{R$^2$} & \textbf{MAE}    & \textbf{RMSE}  & \textbf{R$^2$} & \textbf{MAE}   & \textbf{RMSE}  & \textbf{R$^2$} & \textbf{MAE}   & \textbf{RMSE}  & \textbf{R$^2$} \\ \hline
GAT                   & 0.667          & 0.844          & 0.446          & 1.390           & 1.777          & 0.068          & 2.368          & 2.950          & 0.236          & 0.917          & 1.209          & 0.086          \\
Metapath2vec          & 0.830          & 1.067          & 0.115          & 2.368           & 4.743          & 0.033          & 6.430          & 9.594          & 0.116          & 0.936          & 1.225          & 0.061          \\
MVURE                 & 0.661          & 0.873          & 0.408          & 1.336           & 1.713          & 0.133          & 2.111          & 2.654          & 0.382          & 0.889          & 1.197          & 0.103          \\
MGFN                  & 0.747          & 0.954          & 0.293          & 2.355           & 4.613          & 0.085          & 6.136          & 9.353          & 0.160          & 0.896          & 1.185          & 0.122          \\
HUGAT                 & 0.662          & 0.844          & 0.446          & 1.408           & 1.787          & 0.056          & 2.342          & 2.891          & 0.267          & 0.957          & 1.220          & 0.068          \\
HAFusion              & 0.664          & 0.893          & 0.380          & 1.289           & 1.689          & 0.157          & {\ul 1.840}    & {\ul 2.366}    & {\ul 0.509}    & 0.824          & 1.146          & 0.179          \\
HKGL                  & 0.658          & 0.852          & 0.436          & 1.309           & 1.686          & 0.160          & 1.964          & 2.505          & 0.449          & 0.870          & {\ul 1.140}    & 0.186          \\
HKGL-trans            & 0.698          & 0.936          & 0.319          & 1.381           & 1.752          & 0.094          & 2.055          & 2.598          & 0.408          & 0.891          & 1.167          & {\ul 0.191}    \\ \hline
SLAK-single           & {\ul 0.601}    & {\ul 0.771}    & {\ul 0.538}    & {\ul 1.287}     & {\ul 1.663}    & {\ul 0.183}    & 1.890          & 2.373          & 0.506          & {\ul 0.810}    & {\ul 1.140}    & 0.187          \\
SLAK-comm             & \textbf{0.581} & \textbf{0.752} & \textbf{0.560} & \textbf{1.280}  & \textbf{1.648} & \textbf{0.198} & \textbf{1.839} & \textbf{2.318} & \textbf{0.529} & \textbf{0.797} & \textbf{1.121} & \textbf{0.213} \\ \bottomrule
\end{tabular}
    }
    \label{tbl:main_result_shanghai}
\end{table*}

\subsection{Datasets}
We conduct experiments on two real-world datasets, Beijing and Shanghai, to evaluate our model.\\
    \textbf{Beijing dataset.} It contains 523 regions in Beijing within the Fifth Ring Road, which are partitioned by main road networks. We collect the population data from WorldPop\footnote{https://hub.worldpop.org/geodata/summary?id=24924}. The commercial activity indicators are reflected by the number of firms~\cite{dong2021gridded}. We also collected data from Dianping, one of the most popular review platforms in China. We use the total number of user generated reviews of all businesses to reflect the social activity of a region. The service quality indicator is calculated as the average user rating of all businesses in a region.\\
    \textbf{Shanghai dataset.} It contains 2032 regions in Shanghai, which are also divided by main road networks. The indicators in Shanghai and their data sources are the same as Beijing dataset.
The basic statistics of these datasets are shown in Table~\ref{tbl:datasets_info}.

\subsection{Experiment Settings}
\subsubsection{Baselines}
We compare our model with state-of-the-art graph embedding methods including \textbf{GAT}~\cite{velivckovic2017graph} and \textbf{Metapath2vec}~\cite{dong2017metapath2vec}, and socioeconomic prediction methods including \textbf{MVURE}~\citep{zhang2021multi}, \textbf{HUGAT}~\cite{kim2022effective}, \textbf{MGFN}~\cite{wu2022multi}, \textbf{HAFusion}~\cite{sun2024urban}, and \textbf{HKGL}~\cite{zhou2023hierarchical}. 
We also evaluate a variant of \textbf{HKGL} model named \textbf{HKGL-trans} by transferring region embeddings across tasks. Specifically, we concatenate the task-specific embeddings with embeddings learned from other tasks at the final layer.
The details of these baselines are shown in Appendix~\ref{sec:baseline_details}.
For our proposed SLAK model, we compare two versions of it. \textbf{SLAK-single} denotes the model training on different tasks separately, i.e., the round 1 in Algorithm~\ref{alg:training}. \textbf{SLAK-comm} denotes the model with cross-task communication, i.e., the round 2 in Algorithm~\ref{alg:training}.

\subsubsection{Evaluation Metrics}
We randomly split the regions into train, validation, and test set by 6:2:2, and adopt the commonly used Mean Absolute Error (MAE), Root Mean Square Error (RMSE), and coefficient of determination ($R^2$) as evaluation metrics.

\section{Implementation Details}
\label{sec:implement}
To ensure fair comparison, we fix the dimension of region embeddings to 64 for all models. We search the learning rate from \{0.00001,0.00005,0.0001,0.0005,0.001,0.005\}, batch size from \{64,128\}, and number of R-GCN layers from \{1,2\}. For all baselines, we conduct grid search to find the best hyperparameters. 

\subsection{Overall Performance}
\label{sec:overall_performance}
We present the overall performance of our model and baselines in Table~\ref{tbl:main_result_beijing} and~\ref{tbl:main_result_shanghai}, from which we have the following observations.

First, our model outperforms baseline methods on all eight indicator prediction tasks across two datasets, with improvements in terms of $R^2$ ranging from 4\% to 76\%, demonstrating the effectiveness and robustness of our model.
Specifically, our SLAK-single model is better than baselines on six of the eight indicators, indicating that leveraging LLM agents to extract relevant meta-paths on a single task is promising but not robust enough. Moreover, the cross-task communication mechanism further improves our performance by 3.1\% to 13.9\% in terms of $R^2$, achieving the best results.

Second, HKGL generally achieves the best performance among baselines because it uses UrbanKG to effectively integrate urban data, and defines several sub-KGs to capture diverse knowledge in UrbanKG. However, it still performs worse than our model, which suggests that manually extracting relevant knowledge from urban data may lead to sub-optimal results. On the contrary, we leverage the reasoning and multi-agent collaboration ability of LLM agents to automatically find and refine meta-paths, thus showing considerable performance gain.
We further visualize the prediction results of our model and HKGL in Appendix~\ref{sec:prediction_visualization}.

Third, it is natural to question whether other methods can also benefit from transferring knowledge across tasks. Therefore, we compare with a variant of the best baseline model by concatenating region embeddings learned from other indicators (HKGL vs. HKGL-trans).
We notice that the performance drops on all indicators with embedding transfer. This is because embeddings from different tasks are trained with supervised signals from different indicators, which may not be suitable for a specific task. This suggests that it is non-trivial to transfer knowledge across different tasks through embedding. However, our model shows better performance with cross-task communication mechanism, which further demonstrates the effectiveness of semantic-guided knowledge fusion as well as knowledge sharing through multi-agent collaboration.

\begin{figure*}[t]
    \centering
\hspace{-2mm}
    \subfigure[Beijing]{
    {\label{subfig:ablation_mp_beijing}}
    \includegraphics[width=.24\linewidth]{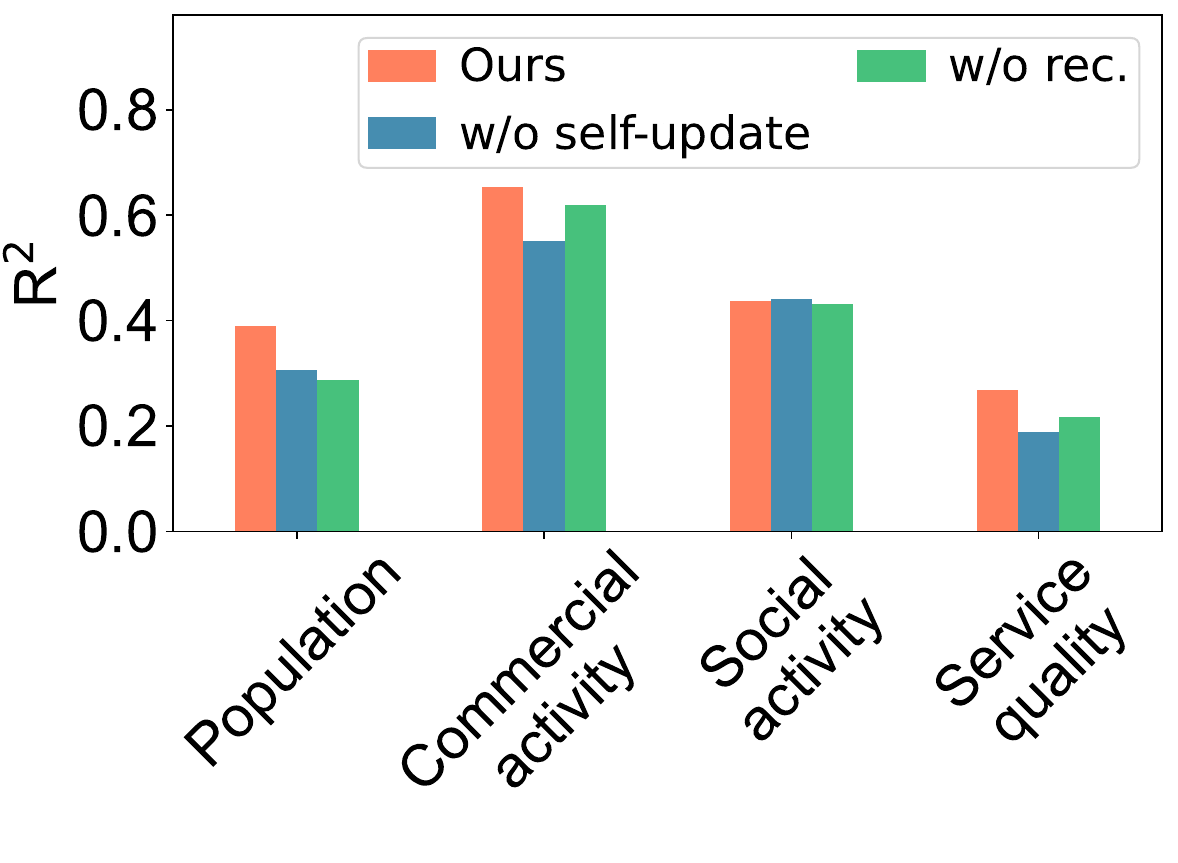}
    }
\hspace{-0mm}
    \subfigure[Shanghai]{
    {\label{subfig:ablation_mp_shanghai}}
    \includegraphics[width=.24\linewidth]{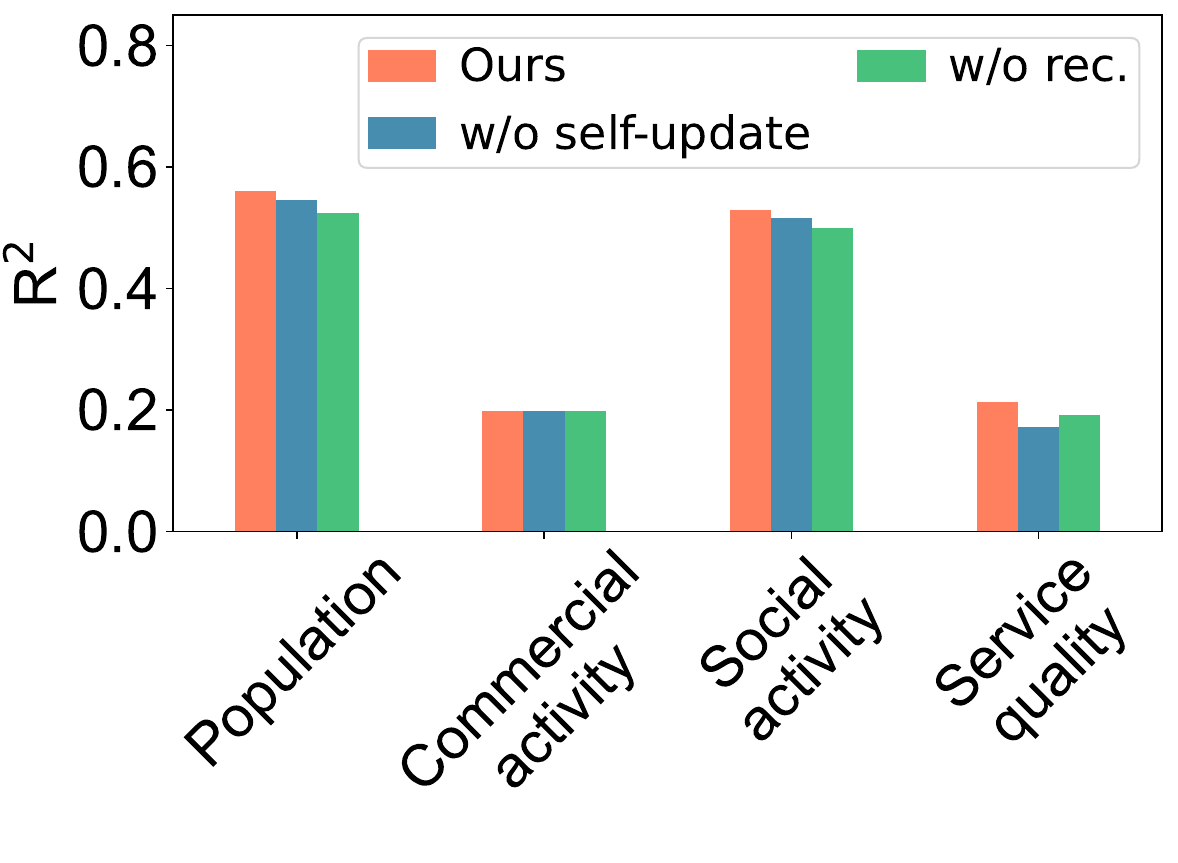}
    }
\hspace{-3mm}
    \subfigure[Beijing]{
    {\label{subfig:ablation_emb_beijing}}
    \includegraphics[width=.24\linewidth]{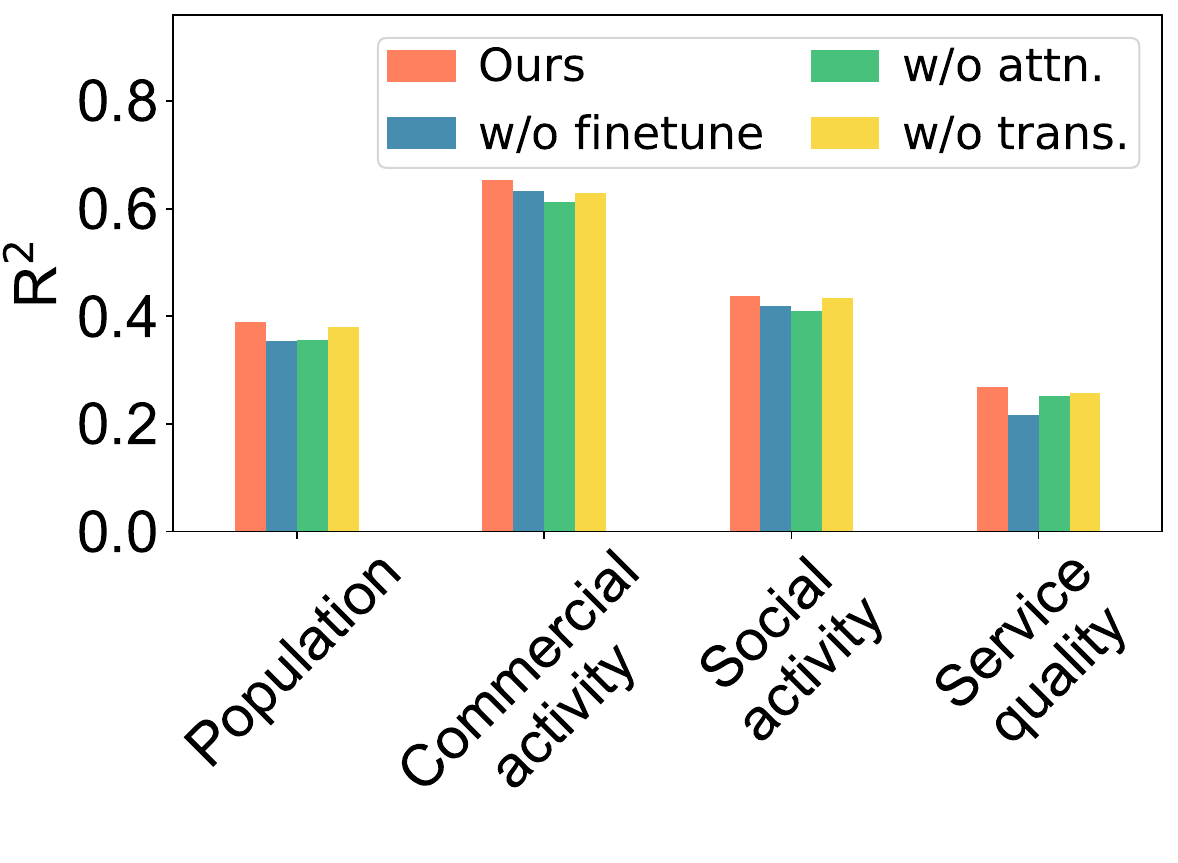}
    }
\hspace{-3mm}
    \subfigure[Shanghai]{
    {\label{subfig:ablation_emb_shanghai}}
    \includegraphics[width=.24\linewidth]{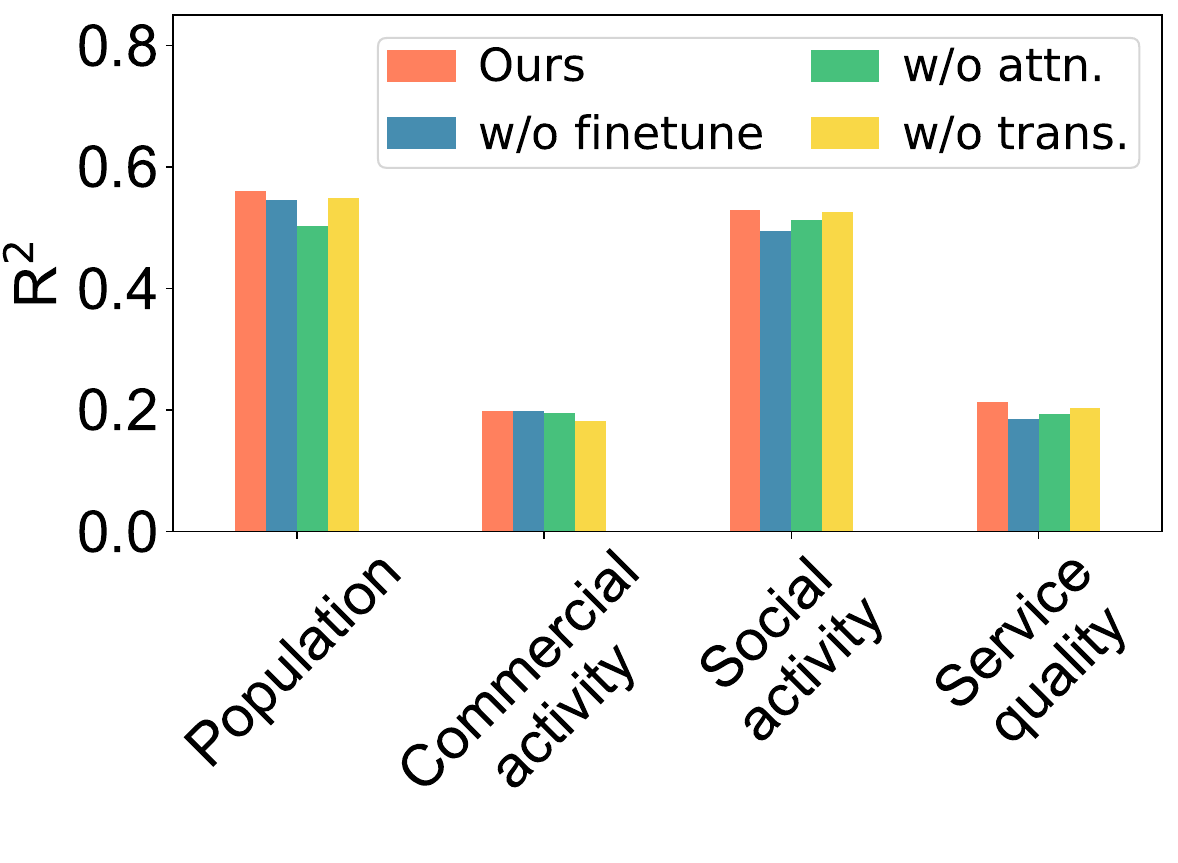}
    }
    \vspace*{-10px}
    \caption{Performance comparison of models without self-update meta-paths (w/o self-update), recommended meta-paths (w/o rec.), embedding LM finetuning (w/o finetune), semantic attention (w/o attn.), and embedding transferred from other tasks (w/o trans.).}
    \label{fig:ablation}
    \vspace*{-5px}
\end{figure*}

\subsection{Ablation Study}


Our framework integrates the reasoning and representation learning paradigms on KG. Therefore, we further conduct ablation studies to demonstrate the effectiveness of such a design. 
We first examine the influence of reasoning on KG, i.e., the meta-path selection through LLM reasoning and multi-agent collaboration.
It has been demonstrated in Section~\ref{sec:overall_performance} that the meta-paths extracted with LLM are better than manually designed meta-paths (the HKGL baseline). Here we test the effect of multi-agent collaboration by removing the meta-path self-update or meta-path recommendation. For example, removing self-update means that we only use the meta-paths recommended by other agents for each task, while removing the recommendation indicates that we only use the self-update meta-paths. As shown in Figure~\ref{fig:ablation}(a-b), the performance drops on most of the indicators across two datasets. Such a result demonstrates the importance of multi-agent collaboration in knowledge sharing across tasks, where different agents find meta-paths from diverse perspectives, enabling more comprehensive knowledge extraction.

Moreover, we evaluate our key designs in the KG representation learning model, i.e., the semantic-enriched entity embedding, the semantic-guided knowledge fusion, and task-aware embedding transfer. 
For entity embedding, we remove the finetuning process of embedding LM, and directly use the GTE model to generate embeddings.
For knowledge fusion, we use a traditional self-attention module to replace our semantic-guided knowledge fusion module in our model.
For the embedding transfer, we omit the embedding fusion module and only use embeddings from the current task for prediction. 
It can be observed in Figure~\ref{fig:ablation}(c-d) that the $R^2$ drops on almost all indicators without these designs, showing that the semantic information of entities and meta-paths in UrbanKG is helpful to enhance the KG representation learning model. Furthermore, the embedding transfer enables knowledge sharing across tasks, leading to better performance.
Overall, the results of ablation studies validate that synergizing LLM agents and KG are crucial for better socioeconomic indicator prediction.

\subsection{Efficiency of Meta-path Optimization}

\begin{table}[htbp!]
    \centering
    \vspace*{-10px}
    \caption{Performance and time cost comparison with meta-path searching algorithms on social activity prediction task. GA and Random represent genetic algorithm and random search, respectively.}
    \vspace*{-10px}
    \resizebox{.95\linewidth}{!}{
\begin{tabular}{c|cccc|cccc}
\toprule
\textbf{}       & \multicolumn{4}{c|}{\textbf{Beijing}}                            & \multicolumn{4}{c}{\textbf{Shanghai}}                            \\
\textbf{Method} & \textbf{MAE}   & \textbf{RMSE}  & \textbf{R$^2$} & \textbf{Time} & \textbf{MAE}   & \textbf{RMSE}  & \textbf{R$^2$} & \textbf{Time} \\ \hline
Random          & 1.799          & 2.543          & 0.414          & 2.5h          & 1.908          & 2.402          & 0.494          & 5.0h          \\
GA              & 1.809          & 2.556          & 0.408          & 2.6h          & 1.918          & 2.391          & 0.498          & 5.1h          \\
Ours            & \textbf{1.657} & \textbf{2.489} & \textbf{0.438} & \textbf{0.2h} & \textbf{1.839} & \textbf{2.318} & \textbf{0.529} & \textbf{0.4h} \\ \bottomrule
\end{tabular}
    }
    \label{tbl:metapath_search}
\end{table}

\begin{table*}[h]
    \centering
    \vspace*{-10px}
    \caption{Best meta-paths found by our model, random search and genetic algorithms on social activity prediction task in Beijing. "Rec by Pop./Com./Ser." indicates the meta-paths recommended by agent of population prediction, commercial activity prediction, and service quality prediction tasks.}
    \vspace*{-5px}
    \resizebox{0.85\textwidth}{!}{
\begin{tabular}{cc|l}
\hline
\multicolumn{2}{c|}{\textbf{Model}}                                & \multicolumn{1}{c}{\textbf{Meta-paths}}                                                                                                                           \\ \hline
\multicolumn{2}{c|}{\multirow{3}{*}{\textbf{SLAK-single}}}         & \textit{$Region\xrightarrow{HasStoreOf}Brand\xrightarrow{ExistIn}POI\xrightarrow{LocateAt}Region$}                                                                \\
\multicolumn{2}{c|}{}                                              & \textit{$Region\xrightarrow{ServedBy}BusinessArea\xrightarrow{Contain}POI\xrightarrow{LocateAt}Region$}                                                           \\
\multicolumn{2}{c|}{}                                              & \textit{$Region\xrightarrow{Has}POI\xrightarrow{HasCategoryOf}Category1\xrightarrow{ExistIn}POI\xrightarrow{LocateAt}Region$}                                     \\ \hline
\multirow{6}{*}{\textbf{SLAK-comm}} & \multirow{3}{*}{Self-update} & \textit{$Region\xrightarrow{Has}POI\xrightarrow{Competitive}POI\xrightarrow{LocateAt}Region$}                                                                     \\
                                    &                              & \textit{$Region\xrightarrow{HasStoreOf}Brand\xrightarrow{BelongTo}Category1\xrightarrow{HasBrandOf}Brand\xrightarrow{ExistIn}POI\xrightarrow{LocateAt}Region$}    \\
                                    &                              & \textit{$Region\xrightarrow{ServedBy}BusinessArea\xrightarrow{Contain}POI\xrightarrow{Competitive}POI\xrightarrow{LocateAt}Region$}                               \\ \cline{2-3} 
                                    & Rec by Pop.                  & \textit{$Region\xrightarrow{PopulationFlowTo}Region\xrightarrow{Has}POI\xrightarrow{HasCategoryOf}Category1\xrightarrow{ExistIn}POI\xrightarrow{LocateAt}Region$} \\ \cline{2-3} 
                                    & Rec by Com.                  & \textit{$Region\xrightarrow{ServedBy}BusinessArea\xrightarrow{Contain}POI\xrightarrow{HasBrandOf}Brand\xrightarrow{ExistIn}POI\xrightarrow{LocateAt}Region$}      \\ \cline{2-3} 
                                    & Rec by Ser.                  & \textit{$Region\xrightarrow{Has}POI\xrightarrow{Competitive}POI\xrightarrow{LocateAt}Region$}                                                                     \\ \hline
\multicolumn{2}{c|}{\multirow{6}{*}{\textbf{Genetic   Algorithm}}} & \textit{$Region\xrightarrow{PopulationFlowTo}Region\xrightarrow{ServedBy}BusinessArea\xrightarrow{Contain}POI\xrightarrow{Competitive}POI$}                       \\
\multicolumn{2}{c|}{}                                              & \textit{$Region\xrightarrow{PopulationFlowTo}Region\xrightarrow{Has}POI\xrightarrow{LocateAt}Region$}                                                             \\
\multicolumn{2}{c|}{}                                              & \textit{$Region\xrightarrow{PopulationInflowFrom}Region\xrightarrow{PopulationInflowFrom}Region\xrightarrow{PopulationFlowTo}Region$}                             \\
\multicolumn{2}{c|}{}                                              & \textit{$Region\xrightarrow{SimilarFunction}Region\xrightarrow{SimilarFunction}Region\xrightarrow{NearBy}Region$}                                                 \\
\multicolumn{2}{c|}{}                                              & \textit{$Region\xrightarrow{PopulationFlowTo}Region\xrightarrow{ServedBy}BusinessArea\xrightarrow{Contain}POI\xrightarrow{LocateAt}Region$}                       \\
\multicolumn{2}{c|}{}                                              & \textit{$Region\xrightarrow{PopulationFlowTo}Region\xrightarrow{HasStoreOf}Brand\xrightarrow{RelatedBrand}Brand$}                                                 \\ \hline
\multicolumn{2}{c|}{\multirow{6}{*}{\textbf{Random   Search}}}     & \textit{$Region\xrightarrow{Has}POI\xrightarrow{HasCategoryOf}Category2\xrightarrow{IsSubCategoryOf}Category1$}                                                   \\
\multicolumn{2}{c|}{}                                              & \textit{$Region\xrightarrow{PopulationFlowTo}Region\xrightarrow{Has}POI\xrightarrow{HasCategoryOf}Category2$}                                                     \\
\multicolumn{2}{c|}{}                                              & \textit{$Region\xrightarrow{SimilarFunction}Region\xrightarrow{HasStoreOf}Brand\xrightarrow{HasPlacedStoreAt}Region$}                                             \\
\multicolumn{2}{c|}{}                                              & \textit{$Region\xrightarrow{ServedBy}BusinessArea\xrightarrow{Serve}Region\xrightarrow{PopulationInflowFrom}Region$}                                              \\
\multicolumn{2}{c|}{}                                              & \textit{$Region\xrightarrow{SimilarFunction}Region\xrightarrow{ServedBy}BusinessArea\xrightarrow{Serve}Region\xrightarrow{PopulationFlowTo}Region$}               \\
\multicolumn{2}{c|}{}                                              & \textit{$Region\xrightarrow{HasStoreOf}Brand\xrightarrow{HasPlacedStoreAt}Region\xrightarrow{PopulationFlowTo}Region$}                                            \\ \hline
\end{tabular}
    }
    \label{tbl:metapaths}
    \vspace*{-10px}
\end{table*}

Here we demonstrate the advantage of our meta-paths extraction method over traditional meta-path searching algorithms.
We compare our model with two meta-path searching algorithms, random search and genetic algorithms. 

In the genetic algorithm, we define a meta-path as a gene, and the chromosome of an individual as a set of 6 meta-paths, which has the same number as our model (3 self-update and 3 recommended meta-paths) for fair comparison. The fitness is calculated as the prediction $R^2$ when using these meta-paths to train our model. We generate 5 individuals in each generation, and calculate the fitness of each. Then we select the best two individuals as parents to perform crossover and mutation to produce the next generation. Specifically, in the crossover operation, one random meta-path is exchanged from the parents. Besides, each meta-path has a probability of 10\% to mutate, i.e., be replaced by a new randomly generated meta-path.
As for the random search algorithm, we also set the number of meta-paths as 6 for each individual, and randomly generate 6 meta-paths with lengths ranging from 2 to 4. Particularly, we start from the entity type \textit{Region}, and sequentially sample a relation with such entity type as the head entity. We repeat the search for 6 iterations, and 5 individuals are generated in each iteration.

The performance and time cost of these algorithms and our model are shown in Table~\ref{tbl:metapath_search}. 
It can be observed that our model achieves better performance and lower time cost. Specifically, our SLAK model outperforms traditional algorithms by 5.8\% and 6.2\% in terms of $R^2$ on Beijing and Shanghai datasets, respectively. This is probably because the number of meta-paths in UrbanKG is large and the potential combination of meta-paths is even significantly larger, making it difficult to find better meta-paths through searching. 
In comparison, the main difference between our method and searching algorithms is that we find meta-paths through reasoning instead of searching. Leveraging the capability of commonsense reasoning, LLM can effectively find meta-paths relevant to the indicator prediction task. Moreover, LLM can optimize the meta-paths through multi-agent collaboration by enabling multiple agents to reason from diverse perspectives, thus leading to better results.
In addition, our model reduces the time cost by over $10\times$, which is because when evaluating each combination of meta-paths, the searching algorithms need to train the KG learning model once, which makes it substantially time-consuming when searching from a large number of individuals. On the contrary, our methods only need to train the KG learning model for a few times. Furthermore, the time cost of LLM communication is much smaller compared with training the model over and over again.

\subsection{Case Study of Discovered Meta-paths}


We present the meta-paths found by our model, random search, and genetic algorithms in Table~\ref{tbl:metapaths}. It can be observed that compared with SLAK-single, the meta-paths discovered in the second round through multi-agent collaboration are generally longer and more diverse. This is probably because these meta-paths are generated by LLM agents based on the previously found ones, which enables agents to consider more complex meta-paths.
Moreover, we find that the meta-paths recommended by other agents are chosen from their own task's perspective. For example, in Beijing dataset, the meta-path recommended by population prediction task $Region\xrightarrow{PopulationFlowTo}Region\xrightarrow{Has}POI\xrightarrow{HasCategoryOf}Category1\xrightarrow{ExistIn}POI\xrightarrow{LocateAt}Region$
 reflect the population flow influence, the one from commercial prediction task $Region\xrightarrow{ServedBy}BusinessArea\xrightarrow{Contain}POI\xrightarrow{HasBrandOf}Brand\xrightarrow{ExistIn}POI\xrightarrow{LocateAt}Region$
 consider the commercial factors between entities like \textit{BusinessArea}, \textit{POI}, and \textit{Brand}. The meta-paths from service quality prediction task $Region\xrightarrow{Has}POI\xrightarrow{Competitive}POI\xrightarrow{LocateAt}Region$
 captures the competitive relationships between POIs in regions, which may impact the service quality as well as social activity. Therefore, the meta-paths found through multi-agent collaboration capture diverse knowledge more comprehensively.
In comparison, the meta-paths generated by searching algorithms are less meaningful and diverse. For example, among the meta-paths generated by the genetic algorithms in Beijing, five out of six meta-paths model the population flow between regions, which may lead to information redundancy.

%% file: 5.related_work.tex
\section{Related Work}
\label{sec:related work}
\subsection{Graph Learning for Socioeconomic Prediction}
Due to the remarkable ability to model non-Euclidean and heterogeneous data, the graph structure and graph learning model have long been utilized for socioeconomic prediction of urban regions. Some studies leverage the mobility data and construct mobility graphs to model the population flow patterns between locations~\cite{wu2022multi,hou2022urban,xu2020attentional}.
Moreover, some studies further consider knowledge in cities from different perspectives through multiple graphs. For instance, HDGE~\cite{wang2017region} constructs a mobility flow graph and spatial graph, and some studies~\cite{zhang2021multi, luo2022urban} propose models to dynamically fuse mobility, POI, and spatial graphs for indicator prediction.

Recently, some studies introduce heterogeneous graph or KG to integrate heterogeneous urban data in a single graph.
HUGAT~\cite{kim2022effective} designs several meta-paths to capture different relationships in urban data from a heterogeneous graph, and aggregate information from meta-path-based neighbors of locations.
Zhou et al. propose a hierarchical KG learning model to learn the global knowledge and domain knowledge from UrbanKG, achieving considerable performance~\cite{zhou2023hierarchical}.
However, existing methods rely on human expertise to identify relevant knowledge from urban data that may help the downstream tasks, which may usually be sub-optimal.

\subsection{Synergy of LLM and KG}
LLMs excel in understanding and generating human language but often suffer from hallucination and a lack of factual accuracy. In contrast, KGs store structured, factual knowledge, which is explicit and interpretable. Therefore, in recent years, researchers have increasingly explored the synergy between LLM and KG to leverage their complementary strengths~\cite{pan2024unifying}. 
Some studies combine LLM with KG for better knowledge representations by fusing the neural networks of language models with KG~\cite{wang2021kepler,zhu2023pre}. Moreover, some studies construct LLM agents that interact with KG for better reasoning. For example, KSL~\cite{feng2023knowledge} lets LLM search on KG to retrieve facts relevant to the questions, StructGPT~\cite{jiang2023structgpt} enables LLM to perform reasoning by traversing on KGs, and Wang et al.~\cite{wang2024knowledge} traverses a knowledge graph across multiple documents to enhance multi-document question answering.
However, existing works that combine LLM and KG only focus on the embedding level or reasoning level. Moreover, there are no previous studies that leverage LLM multi-agent collaboration for knowledge transfer across different tasks. In comparison, we synergizes LLM agent and KG more comprehensively at latent embedding, explicit reasoning and multi-agent collaboration level.

%% file: 6.conclusion.tex
\section{Conclusion}
\label{sec:conclusion}


In this work, we propose a framework that synergizes the ability of LLM agents and KG for better socioeconomic prediction of urban regions. Specifically, we finetune an embedding LM to capture the semantic information in UrbanKG. We then leverage LLM agents to automatically find task-relevant meta-paths, and propose a semantic-guided knowledge fusion model to adaptively fuse diverse knowledge based on semantic information of the meta-paths. Moreover, we enable cross-task knowledge transfer by LLM agent-level collaboration and embedding-level knowledge fusion. Such synergistic model design shows better performance compared with socioeconomic prediction baselines as well as meta-path searching algorithms.
In the future, we plan to consider the dynamic update of UrbanKG and socioeconomic indicators as the urban environment changes over time, and leverage LLM to further capture the temporal patterns of such dynamics.
Another promising direction is to explore the integration of LLM and KG throughout the lifecycle from UrbanKG construction and representation to downstream applications.

%% file: 7.Appendix.tex
\section{Details of UrbanKG}
\label{sec:KG_details}
We present the relations in UrbanKG and their meanings in Table~\ref{tbl:KG_relations}. 
We incorporate three level of POI category, i.e., coarse-grained, medium-grained, and find-grained categories, which are represented as Category1, Category2, and Category3, respectively.

\section{Prompts}
\label{sec:prompt}
\begin{figure*}[h]
    \centering
    \includegraphics[width=.85\linewidth]{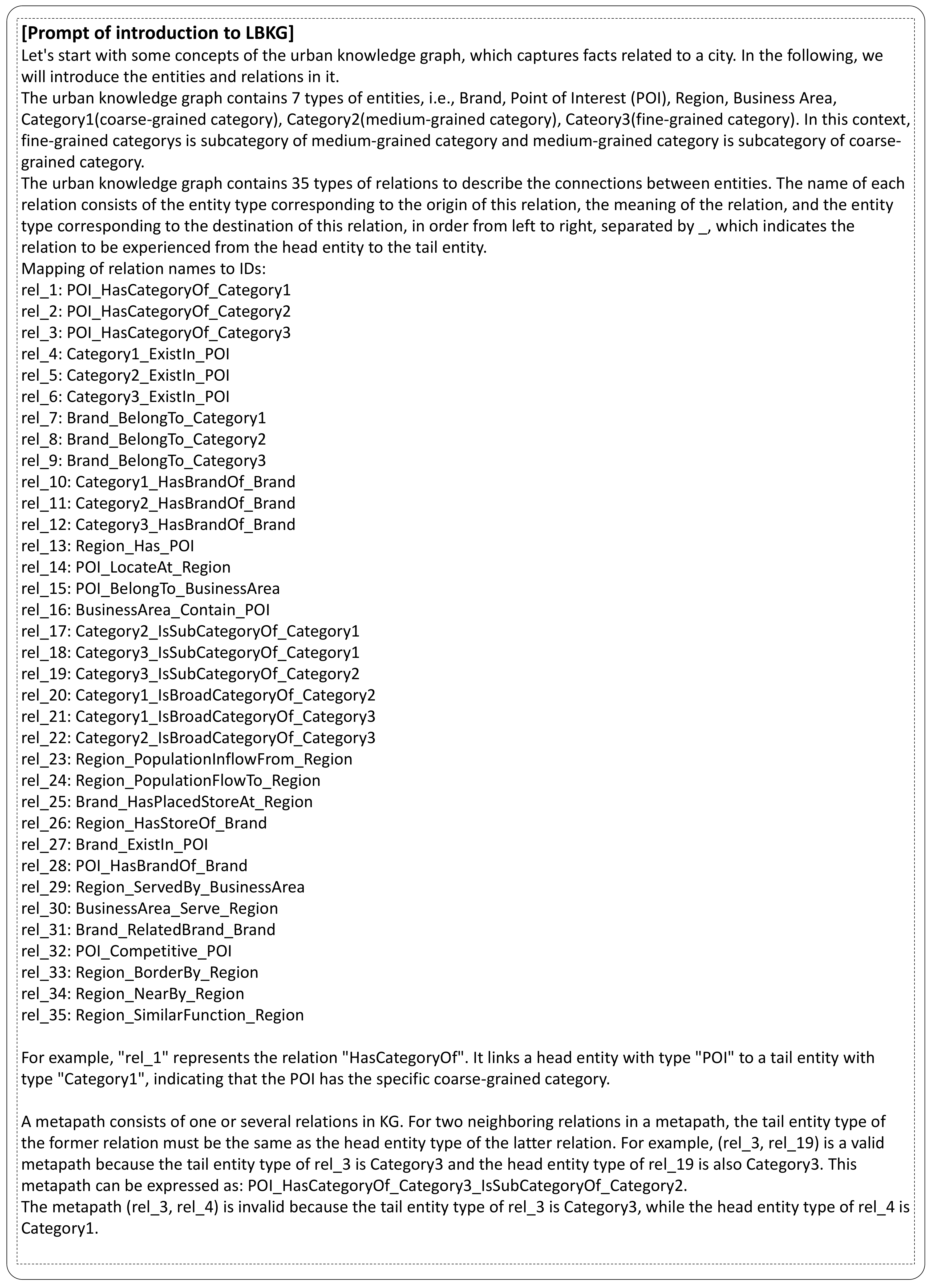}
    \caption{The description of UrbanKG in the prompts.}
    \label{fig:prompt_kginfo}
\end{figure*}

\begin{figure*}[h]
    \centering
    \includegraphics[width=.85\linewidth]{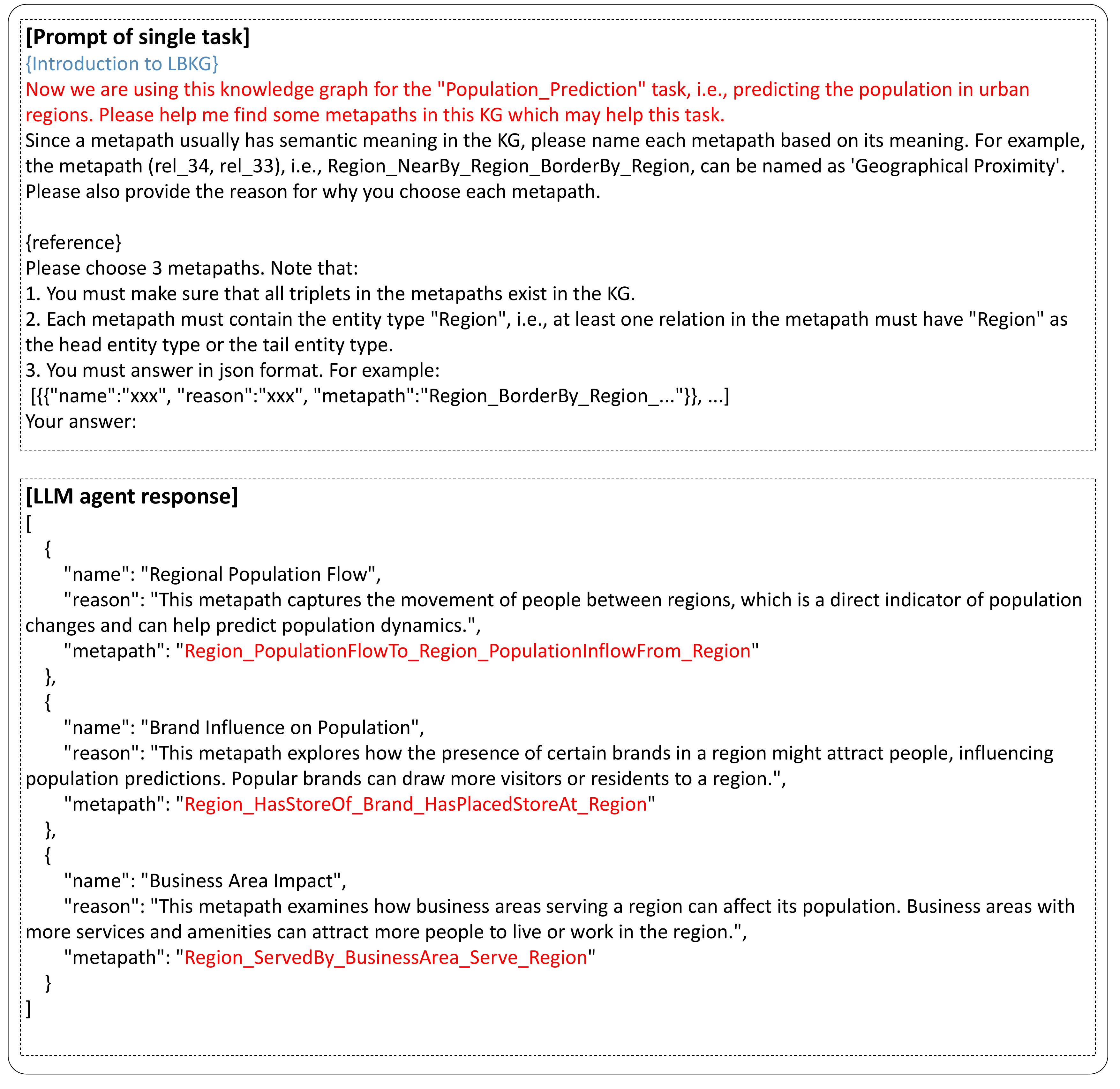}
    \caption{The prompt and response example on single task.}
    \label{fig:prompt_singletask}
\end{figure*}

\begin{figure*}[h]
    \centering
    \includegraphics[width=.85\linewidth]{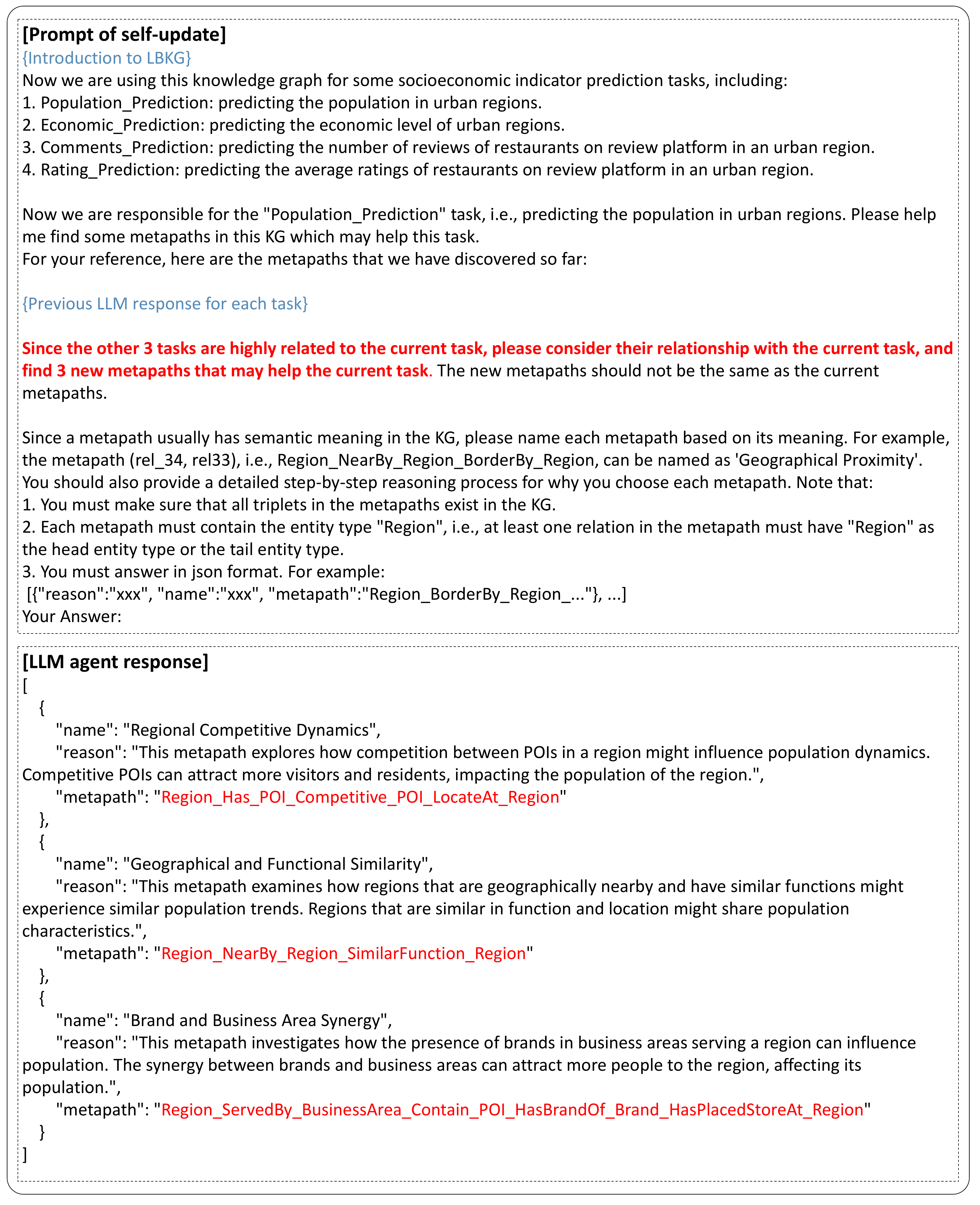}
    \caption{The prompt and response example of meta-paths self-update in cross-task communication.}
    \label{fig:prompt_selfupdate}
\end{figure*}

\begin{figure*}[h]
    \centering
    \includegraphics[width=.85\linewidth]{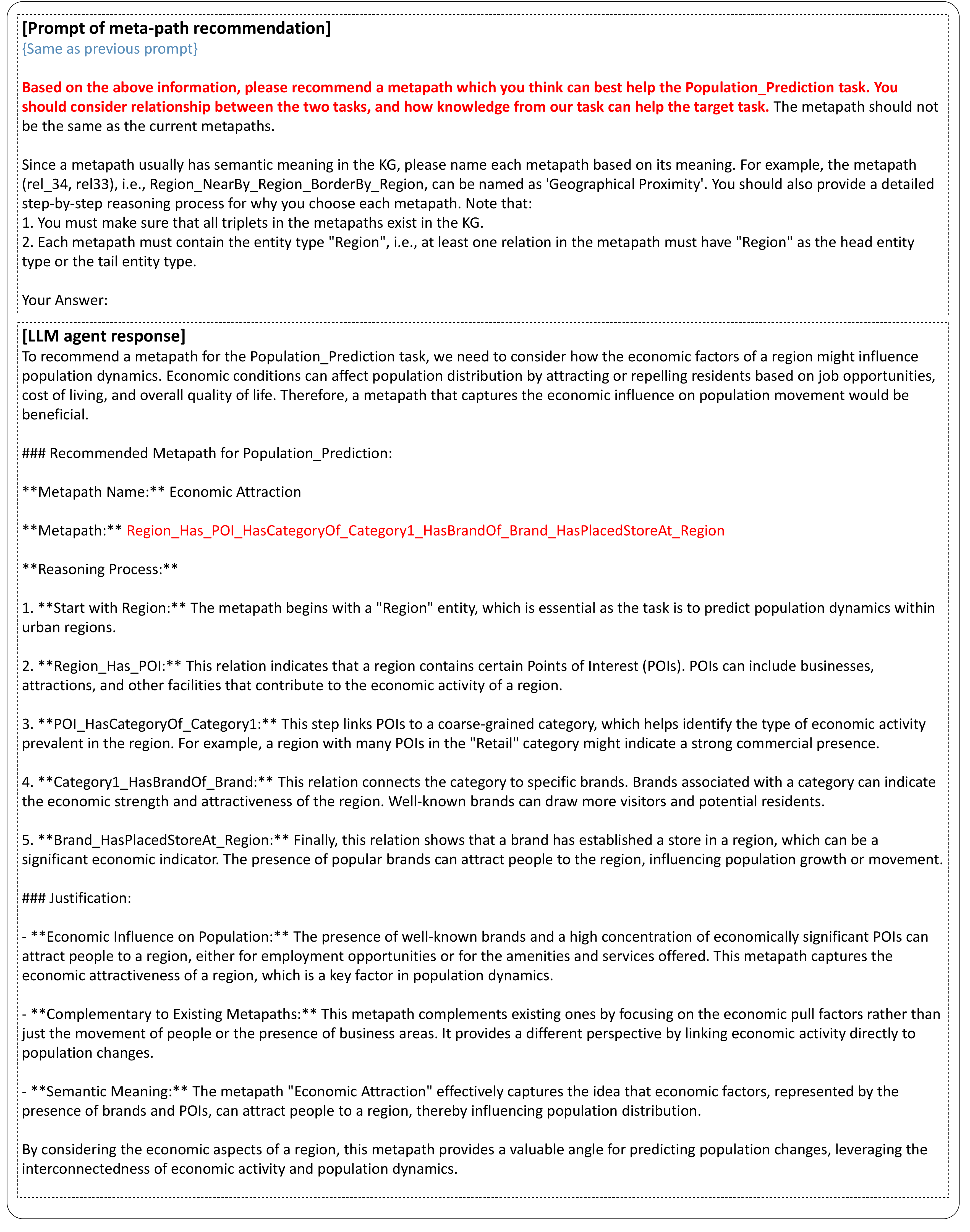}
    \caption{The prompt and response example of meta-paths recommendation in cross-task communication.}
    \label{fig:prompt_rec}
\end{figure*}

We present the detailed prompts and examples of LLM agent response in this Section.
Specifically, we introduce the basic information of UrbanKG to LLM agent as shown in Figure~\ref{fig:prompt_kginfo}.
Figure~\ref{fig:prompt_singletask} shows the examples of prompt and response on single task. Figure~\ref{fig:prompt_selfupdate} and~\ref{fig:prompt_rec} shows the prompt of meta-paths self-update and recommendation in cross-task communication.
We ask the LLM agents to name each meta-path and provide reasons in all prompts.

\section{Details of Baselines}
\label{sec:baseline_details}
We present the details of baselines in this section.
\textbf{Graph embedding methods.}
\begin{itemize}[leftmargin=10px]
    \item \textbf{GAT}~\cite{velivckovic2017graph}: It is a graph convolution network that aggregates information from neighboring nodes with learnable weights.
    \item \textbf{Metapath2vec}~\cite{dong2017metapath2vec}: Since our method is based on meta-paths, we also compare with this model, which learns node embeddings for heterogeneous graph by meta-path-based random walk.
\end{itemize}
\textbf{Socioeconomic prediction methods.}
\begin{itemize}[leftmargin=10px]
    \item \textbf{MVURE}~\citep{zhang2021multi}: This work constructs different graphs to model different types of correlations between regions, and proposes a joint learning module to learn region embeddings. 
    \item \textbf{HUGAT}~\cite{kim2022effective}: It constructs a heterogeneous graph to model urban data, and uses a heterogeneous graph attention network to learn region embeddings for socioeconomic indicator prediction. 
    \item \textbf{MGFN}~\cite{wu2022multi}: It constructs mobility graphs and leverages multi-level attention mechanism to learn region embeddings. 
    \item \textbf{HAFusion}~\cite{sun2024urban}: It learns high-order correlations between different regions and different region features with a dual-feature attentive fusion module. 
    \item \textbf{HKGL}~\cite{zhou2023hierarchical}: It uses a hierarchical KG learning model to learn global and domain knowledge from UrbanKG. We also evaluate a variant of this model named \textbf{HKGL-trans} by transferring region embeddings across tasks. Specifically, we concatenate the task-specific embeddings with embeddings learned from other tasks at the final layer.
\end{itemize}

\section{Visualization of Prediction Results}
\label{sec:prediction_visualization}
\begin{figure}[h]
    \centering
    \includegraphics[width=.9\linewidth]{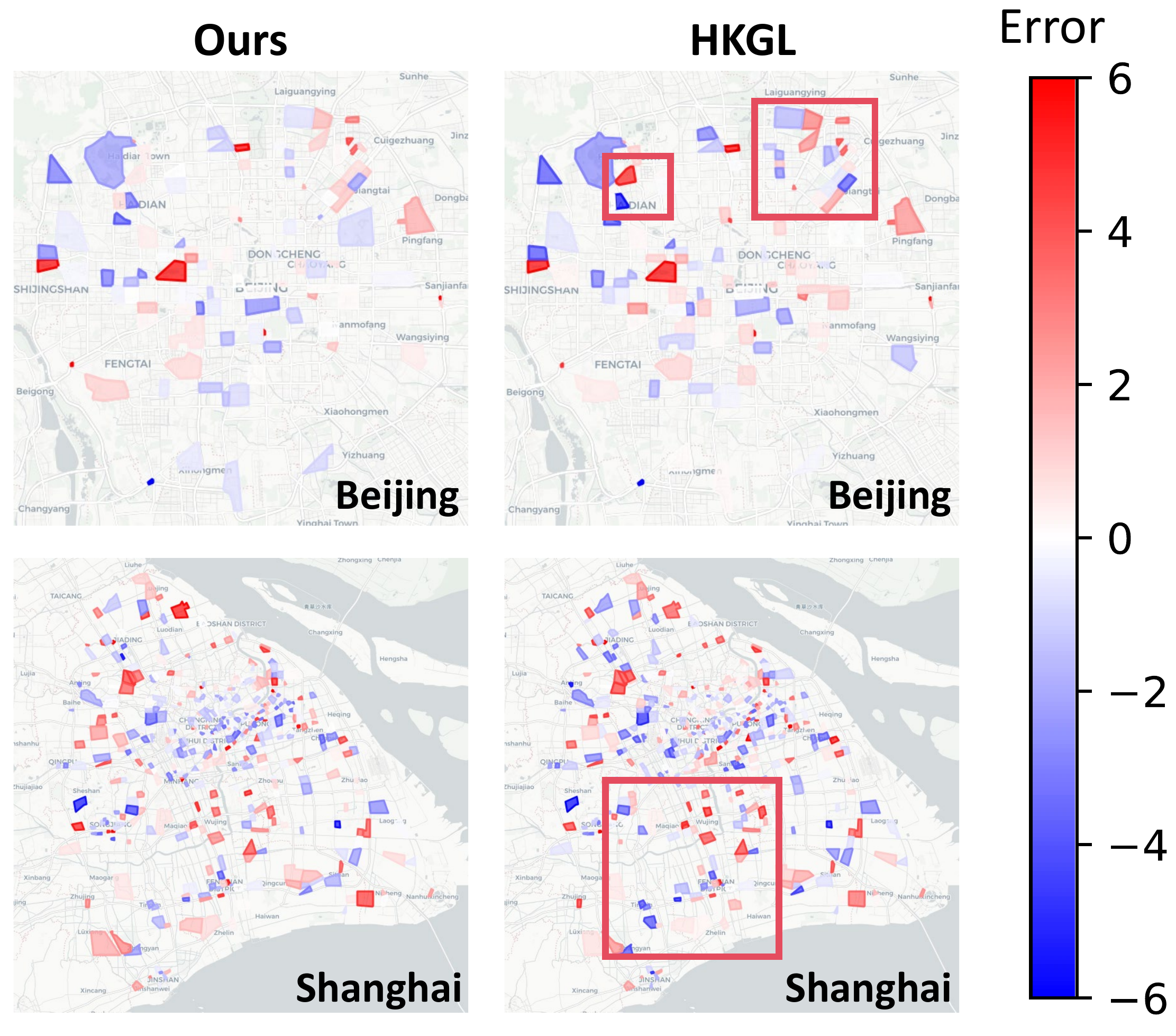}
    \caption{Visualization of prediction error of social activity (measured in log scale). We use red squares to mark the regions where baseline performs worse.}
    \label{fig:error_visualization}
\end{figure}

To intuitively understand the socioeconomic prediction result, we visualize the indicator prediction error on test regions of our model and the best baseline model on social activity prediction task. As shown in Figure~\ref{fig:error_visualization}, our model generally shows smaller errors compared with HKGL. Specifically, our model shows better accuracy in Haidian and Wangjing areas, which are known for their high density of tech companies as well as vibrant social activities. In addition, our model also performs better in Fengxian district in southwestern Shanghai. Such results further demonstrate the accuracy of our model in socioeconomic indicator prediction tasks.

\newpage
\begin{table*}[h]
    \centering
    \caption{The details of relations in UrbanKG.}
    \resizebox{0.8\textwidth}{!}{
\begin{tabular}{c|c|c}
\hline
\textbf{Relation}                & \textbf{Head \& Tail   Entity Types} & \textbf{Meaning}                               \\ \hline
\textit{HasCategory1Of}          & (POI, Category1)                     & Coarse-grained category of POI                 \\
\textit{HasCategory2Of}          & (POI, Category2)                     & Medium-grained category of POI                 \\
\textit{HasCategory3Of}          & (POI, Category3)                     & Fine-grained category of POI                   \\
\textit{Category1ExistIn}        & (Category1, POI)                     & Coarse-grained category contains POI           \\
\textit{Category2ExistIn}        & (Category2, POI)                     & Medium-grained category contains POI           \\
\textit{Category3ExistIn}        & (Category3, POI)                     & Fine-grained category contains POI             \\
\textit{BelongToCategory1}       & (Brand, Category1)                   & Coarse-grained category of Brand               \\
\textit{BelongToCategory2}       & (Brand, Category2)                   & Medium-grained category of Brand               \\
\textit{BelongToCategory3}       & (Brand, Category3)                   & Fine-grained category of Brand                 \\
\textit{Category1HasBrandOf}     & (Category1, Brand)                   & Coarse-grained category contains Brand         \\
\textit{Category2HasBrandOf}     & (Category2, Brand)                   & Medium-grained category contains Brand         \\
\textit{Category3HasBrandOf}     & (Category3, Brand)                   & Fine-grained category contains Brand           \\
\textit{Has}                     & (Region, POI)                        & Region contains POI                            \\
\textit{LocateAt}                & (POI, Region)                        & POI locates at the region                      \\
\textit{BelongTo}                & (POI, BusinessArea)                  & POI locates at the BusinessArea                \\
\textit{Contain}                 & (BusinessArea, POI)                  & BusinessArea contains POI                      \\
\textit{IsSubCategoryOf\_2to1}   & (Category2, Category1)               & Category2 is a sub-category of Category1       \\
\textit{IsSubCategoryOf\_3to1}   & (Category3, Category1)               & Category3 is a sub-category of Category1       \\
\textit{IsSubCategoryOf\_3to2}   & (Category3, Category2)               & Category3 is a sub-category of Category2       \\
\textit{IsBroadCategoryOf\_1to2} & (Category1, Category2)               & Category1 is a broad-category of Category2     \\
\textit{IsBroadCategoryOf\_1to3} & (Category1, Category3)               & Category1 is a broad-category of Category3     \\
\textit{IsBroadCategoryOf\_2to3} & (Category2, Category3)               & Category2 is a broad-category of Category3     \\
\textit{PopulationInflowFrom}    & (Region, Region)                     & Region has a large population flow to Region   \\
\textit{PopulationFlowTo}        & (Region, Region)                     & Region has a large population flow from Region \\
\textit{HasPlacedStoreAt}        & (Brand, Region)                      & Brand has opened store at Region               \\
\textit{HasStoreOf}              & (Region, Brand)                      & Region contains POI of the Brand               \\
\textit{BrandExistIn}            & (Brand, POI)                         & Brand contains POI                             \\
\textit{HasBrandOf}              & (POI, Brand)                         & Brand of POI                                   \\
\textit{ServedBy}                & (Region, BusinessArea)               & Region covered by BusinessArea                 \\
\textit{Serve}                   & (BusinessArea, Region)               & BusinessArea covers the region                 \\
\textit{RelatedBrand}            & (Brand, Brand)                       & Brands related to each other                   \\
\textit{Competitive}             & (POI, POI)                           & Nearby POIs with the same brand                \\
\textit{BorderBy}                & (Region, Region)                     & Regions share part of the boundary             \\
\textit{NearBy}                  & (Region, Region)                     & Regions lie within a certain distance          \\
\textit{SimilarFunction}         & (Region, Region)                     & Regions with similar POI distribution          \\ \hline
\end{tabular}
    }
    \label{tbl:KG_relations}
\end{table*}

%% file: ref.bib
@inproceedings{zhou2023hierarchical,
  title={Hierarchical knowledge graph learning enabled socioeconomic indicator prediction in location-based social network},
  author={Zhou, Zhilun and Liu, Yu and Ding, Jingtao and Jin, Depeng and Li, Yong},
  booktitle={Proceedings of the ACM Web Conference 2023},
  pages={122--132},
  year={2023}
}

@article{liu2023urbankg,
  title={Urbankg: An urban knowledge graph system},
  author={Liu, Yu and Ding, Jingtao and Fu, Yanjie and Li, Yong},
  journal={ACM Transactions on Intelligent Systems and Technology},
  volume={14},
  number={4},
  pages={1--25},
  year={2023},
  publisher={ACM New York, NY}
}

@inproceedings{liu2023knowsite,
  title={KnowSite: Leveraging Urban Knowledge Graph for Site Selection},
  author={Liu, Yu and Ding, Jingtao and Li, Yong},
  booktitle={Proceedings of the 31st ACM International Conference on Advances in Geographic Information Systems},
  pages={1--12},
  year={2023}
}

@article{zhao2024large,
  title={Large language models as commonsense knowledge for large-scale task planning},
  author={Zhao, Zirui and Lee, Wee Sun and Hsu, David},
  journal={Advances in Neural Information Processing Systems},
  volume={36},
  year={2024}
}

@inproceedings{xiao2023chain,
  title={Chain-of-Experts: When LLMs Meet Complex Operations Research Problems},
  author={Xiao, Ziyang and Zhang, Dongxiang and Wu, Yangjun and Xu, Lilin and Wang, Yuan Jessica and Han, Xiongwei and Fu, Xiaojin and Zhong, Tao and Zeng, Jia and Song, Mingli and others},
  booktitle={The Twelfth International Conference on Learning Representations},
  year={2023}
}

@inproceedings{li2023camel,
    title={{CAMEL}: Communicative Agents for ''Mind'' Exploration of Large Language Model Society},
    author={Guohao Li and Hasan Abed Al Kader Hammoud and Hani Itani and Dmitrii Khizbullin and Bernard Ghanem},
    booktitle={Thirty-seventh Conference on Neural Information Processing Systems},
    year={2023},
    url={https://openreview.net/forum?id=3IyL2XWDkG}
}

@misc{hong2023metagpt,
    title={MetaGPT: Meta Programming for A Multi-Agent Collaborative Framework}, 
    author={Sirui Hong and Mingchen Zhuge and Jonathan Chen and Xiawu Zheng and Yuheng Cheng and Ceyao Zhang and Jinlin Wang and Zili Wang and Steven Ka Shing Yau and Zijuan Lin and Liyang Zhou and Chenyu Ran and Lingfeng Xiao and Chenglin Wu and Jürgen Schmidhuber},
    year={2023},
    eprint={2308.00352},
    archivePrefix={arXiv},
    primaryClass={cs.AI}
}

@book{sun2012mining,
  title={Mining heterogeneous information networks: principles and methodologies},
  author={Sun, Yizhou and Han, Jiawei},
  year={2012},
  publisher={Morgan \& Claypool Publishers}
}

@inproceedings{chen2024large,
  title={Large Language Model-driven Meta-structure Discovery in Heterogeneous Information Network},
  author={Chen, Lin and Xu, Fengli and Li, Nian and Han, Zhenyu and Wang, Meng and Li, Yong and Hui, Pan},
  booktitle={Proceedings of the 30th ACM SIGKDD Conference on Knowledge Discovery and Data Mining},
  pages={307--318},
  year={2024}
}

@inproceedings{dong2017metapath2vec,
  title={metapath2vec: Scalable representation learning for heterogeneous networks},
  author={Dong, Yuxiao and Chawla, Nitesh V and Swami, Ananthram},
  booktitle={Proceedings of the 23rd ACM SIGKDD international conference on knowledge discovery and data mining},
  pages={135--144},
  year={2017}
}

@article{li2023towards,
  title={Towards general text embeddings with multi-stage contrastive learning},
  author={Li, Zehan and Zhang, Xin and Zhang, Yanzhao and Long, Dingkun and Xie, Pengjun and Zhang, Meishan},
  journal={arXiv preprint arXiv:2308.03281},
  year={2023}
}

@article{muennighoff2022mteb,
  title={MTEB: Massive text embedding benchmark},
  author={Muennighoff, Niklas and Tazi, Nouamane and Magne, Lo{\"\i}c and Reimers, Nils},
  journal={arXiv preprint arXiv:2210.07316},
  year={2022}
}

@inproceedings{lan2024stance,
  title={Stance detection with collaborative role-infused llm-based agents},
  author={Lan, Xiaochong and Gao, Chen and Jin, Depeng and Li, Yong},
  booktitle={Proceedings of the International AAAI Conference on Web and Social Media},
  volume={18},
  pages={891--903},
  year={2024}
}

@article{chan2023chateval,
  title={Chateval: Towards better llm-based evaluators through multi-agent debate},
  author={Chan, Chi-Min and Chen, Weize and Su, Yusheng and Yu, Jianxuan and Xue, Wei and Zhang, Shanghang and Fu, Jie and Liu, Zhiyuan},
  journal={arXiv preprint arXiv:2308.07201},
  year={2023}
}

@article{wei2022chain,
  title={Chain-of-thought prompting elicits reasoning in large language models},
  author={Wei, Jason and Wang, Xuezhi and Schuurmans, Dale and Bosma, Maarten and Xia, Fei and Chi, Ed and Le, Quoc V and Zhou, Denny and others},
  journal={Advances in neural information processing systems},
  volume={35},
  pages={24824--24837},
  year={2022}
}

@article{pan2024unifying,
  title={Unifying large language models and knowledge graphs: A roadmap},
  author={Pan, Shirui and Luo, Linhao and Wang, Yufei and Chen, Chen and Wang, Jiapu and Wu, Xindong},
  journal={IEEE Transactions on Knowledge and Data Engineering},
  year={2024},
  publisher={IEEE}
}

@article{wang2021kepler,
  title={KEPLER: A unified model for knowledge embedding and pre-trained language representation},
  author={Wang, Xiaozhi and Gao, Tianyu and Zhu, Zhaocheng and Zhang, Zhengyan and Liu, Zhiyuan and Li, Juanzi and Tang, Jian},
  journal={Transactions of the Association for Computational Linguistics},
  volume={9},
  pages={176--194},
  year={2021},
  publisher={MIT Press One Rogers Street, Cambridge, MA 02142-1209, USA journals-info~…}
}

@article{zhu2023pre,
  title={Pre-training language model incorporating domain-specific heterogeneous knowledge into a unified representation},
  author={Zhu, Hongyin and Peng, Hao and Lyu, Zhiheng and Hou, Lei and Li, Juanzi and Xiao, Jinghui},
  journal={Expert Systems with Applications},
  volume={215},
  pages={119369},
  year={2023},
  publisher={Elsevier}
}

@inproceedings{wang2024knowledge,
  title={Knowledge graph prompting for multi-document question answering},
  author={Wang, Yu and Lipka, Nedim and Rossi, Ryan A and Siu, Alexa and Zhang, Ruiyi and Derr, Tyler},
  booktitle={Proceedings of the AAAI Conference on Artificial Intelligence},
  volume={38},
  number={17},
  pages={19206--19214},
  year={2024}
}

@article{jiang2023structgpt,
  title={Structgpt: A general framework for large language model to reason over structured data},
  author={Jiang, Jinhao and Zhou, Kun and Dong, Zican and Ye, Keming and Zhao, Wayne Xin and Wen, Ji-Rong},
  journal={arXiv preprint arXiv:2305.09645},
  year={2023}
}

@article{feng2023knowledge,
  title={Knowledge solver: Teaching llms to search for domain knowledge from knowledge graphs},
  author={Feng, Chao and Zhang, Xinyu and Fei, Zichu},
  journal={arXiv preprint arXiv:2309.03118},
  year={2023}
}

@inproceedings{wang2016crime,
  title={Crime rate inference with big data},
  author={Wang, Hongjian and Kifer, Daniel and Graif, Corina and Li, Zhenhui},
  booktitle={Proceedings of the 22nd ACM SIGKDD International Conference on Knowledge Discovery and Data Mining},
  pages={635--644},
  year={2016}
}

@article{hou2022urban,
  title={Urban Region Profiling With Spatio-Temporal Graph Neural Networks},
  author={Hou, Mingliang and Xia, Feng and Gao, Haoran and Chen, Xin and Chen, Honglong},
  journal={IEEE Transactions on Computational Social Systems},
  year={2022},
  publisher={IEEE}
}

@article{luo2022urban,
  title={Urban Region Profiling via A Multi-Graph Representation Learning Framework},
  author={Luo, Yan and Chung, Fu-lai and Chen, Kai},
  journal={arXiv preprint arXiv:2202.02074},
  year={2022}
}

@article{wu2022multi,
  title={Multi-Graph Fusion Networks for Urban Region Embedding},
  author={Wu, Shangbin and Yan, Xu and Fan, Xiaoliang and Pan, Shirui and Zhu, Shichao and Zheng, Chuanpan and Cheng, Ming and Wang, Cheng},
  journal={arXiv preprint arXiv:2201.09760},
  year={2022}
}

@article{kim2022effective,
  title={Effective Urban Region Representation Learning Using Heterogeneous Urban Graph Attention Network (HUGAT)},
  author={Kim, Namwoo and Yoon, Yoonjin},
  journal={arXiv preprint arXiv:2202.09021},
  year={2022}
}

@inproceedings{yao2018representing,
  title={Representing urban functions through zone embedding with human mobility patterns},
  author={Yao, Zijun and Fu, Yanjie and Liu, Bin and Hu, Wangsu and Xiong, Hui},
  booktitle={Proceedings of the Twenty-Seventh International Joint Conference on Artificial Intelligence (IJCAI-18)},
  year={2018}
}

@inproceedings{zhang2021multi,
  title={Multi-view joint graph representation learning for urban region embedding},
  author={Zhang, Mingyang and Li, Tong and Li, Yong and Hui, Pan},
  booktitle={Proceedings of the Twenty-Ninth International Conference on International Joint Conferences on Artificial Intelligence},
  pages={4431--4437},
  year={2021}
}

@inproceedings{wang2017region,
  title={Region representation learning via mobility flow},
  author={Wang, Hongjian and Li, Zhenhui},
  booktitle={Proceedings of the 2017 ACM on Conference on Information and Knowledge Management},
  pages={237--246},
  year={2017}
}

@article{yang2017predicting,
  title={Predicting commercial activeness over urban big data},
  author={Yang, Su and Wang, Minjie and Wang, Wenshan and Sun, Yi and Gao, Jun and Zhang, Weishan and Zhang, Jiulong},
  journal={Proceedings of the ACM on Interactive, Mobile, Wearable and Ubiquitous Technologies},
  volume={1},
  number={3},
  pages={1--20},
  year={2017},
  publisher={ACM New York, NY, USA}
}

@article{dong2019predicting,
  title={Predicting neighborhoods’ socioeconomic attributes using restaurant data},
  author={Dong, Lei and Ratti, Carlo and Zheng, Siqi},
  journal={Proceedings of the National Academy of Sciences},
  volume={116},
  number={31},
  pages={15447--15452},
  year={2019},
  publisher={National Acad Sciences}
}

@inproceedings{xu2020attentional,
  title={Attentional multi-graph convolutional network for regional economy prediction with open migration data},
  author={Xu, Fengli and Li, Yong and Xu, Shusheng},
  booktitle={Proceedings of the 26th ACM SIGKDD International Conference on Knowledge Discovery \& Data Mining},
  pages={2225--2233},
  year={2020}
}

@inproceedings{schlichtkrull2018modeling,
  title={Modeling relational data with graph convolutional networks},
  author={Schlichtkrull, Michael and Kipf, Thomas N and Bloem, Peter and Berg, Rianne van den and Titov, Ivan and Welling, Max},
  booktitle={European Semantic Web Conference},
  pages={593--607},
  year={2018},
  organization={Springer}
}

@article{velivckovic2017graph,
  title={Graph attention networks},
  author={Veli{\v{c}}kovi{\'c}, Petar and Cucurull, Guillem and Casanova, Arantxa and Romero, Adriana and Lio, Pietro and Bengio, Yoshua},
  journal={arXiv preprint arXiv:1710.10903},
  year={2017}
}

@article{dong2021gridded,
  title={A gridded establishment dataset as a proxy for economic activity in China},
  author={Dong, Lei and Yuan, Xiaohui and Li, Meng and Ratti, Carlo and Liu, Yu},
  journal={Scientific Data},
  volume={8},
  number={1},
  pages={1--9},
  year={2021},
  publisher={Nature Publishing Group}
}

@article{wang2021spatio,
  title={Spatio-Temporal Urban Knowledge Graph Enabled Mobility Prediction},
  author={Wang, Huandong and Yu, Qiaohong and Liu, Yu and Jin, Depeng and Li, Yong},
  journal={Proceedings of the ACM on Interactive, Mobile, Wearable and Ubiquitous Technologies},
  volume={5},
  number={4},
  pages={1--24},
  year={2021},
  publisher={ACM New York, NY, USA}
}

@article{liu2021improving,
  title={Improving Location Recommendation with Urban Knowledge Graph},
  author={Liu, Chang and Gao, Chen and Jin, Depeng and Li, Yong},
  journal={arXiv preprint arXiv:2111.01013},
  year={2021}
}

@article{bordes2013translating,
  title={Translating embeddings for modeling multi-relational data},
  author={Bordes, Antoine and Usunier, Nicolas and Garcia-Duran, Alberto and Weston, Jason and Yakhnenko, Oksana},
  journal={Advances in neural information processing systems},
  volume={26},
  year={2013}
}

@inproceedings{sun2024urban,
  title={Urban region representation learning with attentive fusion},
  author={Sun, Fengze and Qi, Jianzhong and Chang, Yanchuan and Fan, Xiaoliang and Karunasekera, Shanika and Tanin, Egemen},
  booktitle={2024 IEEE 40th International Conference on Data Engineering (ICDE)},
  pages={4409--4421},
  year={2024},
  organization={IEEE}
}

@inproceedings{wang2023reasoning,
  title={Reasoning through memorization: Nearest neighbor knowledge graph embeddings},
  author={Wang, Peng and Xie, Xin and Wang, Xiaohan and Zhang, Ninyu},
  booktitle={CCF International Conference on Natural Language Processing and Chinese Computing},
  pages={111--122},
  year={2023},
  organization={Springer}
}

@article{wang2022language,
  title={Language models as knowledge embeddings},
  author={Wang, Xintao and He, Qianyu and Liang, Jiaqing and Xiao, Yanghua},
  journal={arXiv preprint arXiv:2206.12617},
  year={2022}
}

@inproceedings{xie2023lambdakg,
  title={LambdaKG: A Library for Pre-trained Language Model-Based Knowledge Graph Embeddings},
  author={Xie, Xin and Li, Zhoubo and Wang, Xiaohan and Xi, ZeKun and Zhang, Ningyu},
  booktitle={Proceedings of the 13th International Joint Conference on Natural Language Processing and the 3rd Conference of the Asia-Pacific Chapter of the Association for Computational Linguistics: System Demonstrations},
  pages={25--33},
  year={2023}
}
